\UseRawInputEncoding
\documentclass{article} % For LaTeX2e
\usepackage{iclr2027_conference,times}

\usepackage{amsmath,amsfonts,bm}

\def\eqref#1{equation~\ref{#1}}
\def\1{\bm{1}}

\DeclareMathAlphabet{\mathsfit}{\encodingdefault}{\sfdefault}{m}{sl}
\SetMathAlphabet{\mathsfit}{bold}{\encodingdefault}{\sfdefault}{bx}{n}

\usepackage{hyperref}
\usepackage{url}
\usepackage{lipsum}
\usepackage{tikzducks}
\usepackage{graphicx}
\usepackage[labelfont=bf]{caption}
\usepackage{enumitem}
\usepackage{titletoc}
\usepackage{wrapfig}
\usepackage{booktabs}
\usepackage{multirow}
\usepackage{caption}
\usepackage[table]{xcolor}
\usepackage{xcolor}
\usepackage{tcolorbox}
\usepackage{bbm}
\usepackage{amsmath}
\usepackage{algorithm}
\usepackage{algpseudocode}
\usepackage{amssymb}
\usepackage{mathrsfs}

\title{\ours{}: 3D Scene Generation with Objects\\ That Fit Together}

\usepackage{hyperref}

\title{\ours{}: 3D Scene Generation with Objects\\ That Fit Together}

\iclrfinalcopy

\author{%
Jaeyeong Kim \quad
Jinhyuk Jang \quad
Jongmin Lee \quad
Kyehong Park \quad
Seungryong Kim\textsuperscript{\(\dagger\)}\\[2mm]
KAIST AI\\[1mm]
Project page: \url{https://cvlab-kaist.github.io/Tetris3D}%
}

\newcommand{\ours}{Tetris3D}

\newcommand{\appendixtoc}{%
  \begingroup
  \titlecontents{section}
    [2.5em]                         % left indent
    {\addvspace{0.7em}\bfseries}  % style before entry
    {\contentslabel{2.5em}}       % numbered entry
    {}                            % unnumbered entry
    {}                            % no page number
  \titlecontents{subsection}
    [6em]
    {}
    {\contentslabel{3.2em}}
    {}
    {}
  \printcontents[appendix]{}{1}{%
    \setcounter{tocdepth}{2}%
  }
  \endgroup
}

\tcbuselibrary{listings,breakable,skins}

\newtcblisting[auto counter]{promptbox}[2][]{
    enhanced,
    breakable,
    listing only,
    title={Prompt~\thetcbcounter: #2},
    fonttitle=\bfseries,
    coltitle=black,
    colbacktitle=gray!15,
    colback=gray!3,
    colframe=gray!60,
    boxrule=0.5pt,
    arc=2pt,
    left=2mm,
    right=2mm,
    listing options={
        language={},
        basicstyle=\ttfamily\scriptsize,
        breaklines=true,
        columns=fullflexible,
        keepspaces=true,
        showstringspaces=false,
        numbers=none
    },
    #1
}

\definecolor{shadegreen}{RGB}{215,233,217}
\definecolor{shadebeige}{RGB}{242,230,201}
\definecolor{shadepeach}{RGB}{240,217,206}

\begin{document}

\maketitle

% Corresponding-author footnote; 2 corresponds to the dagger symbol.
\begingroup
\renewcommand{\thefootnote}{\fnsymbol{footnote}}
\footnotetext[2]{Corresponding author.}
\endgroup

\vspace{-2em}
\begin{figure}[h]
    \centering
    % \resizebox{\columnwidth}{8cm}{
        % \begin{tikzpicture}
        %     \duck[water=cyan, sunglasses=black]
        % \end{tikzpicture}
    % }
\includegraphics[width=\linewidth]{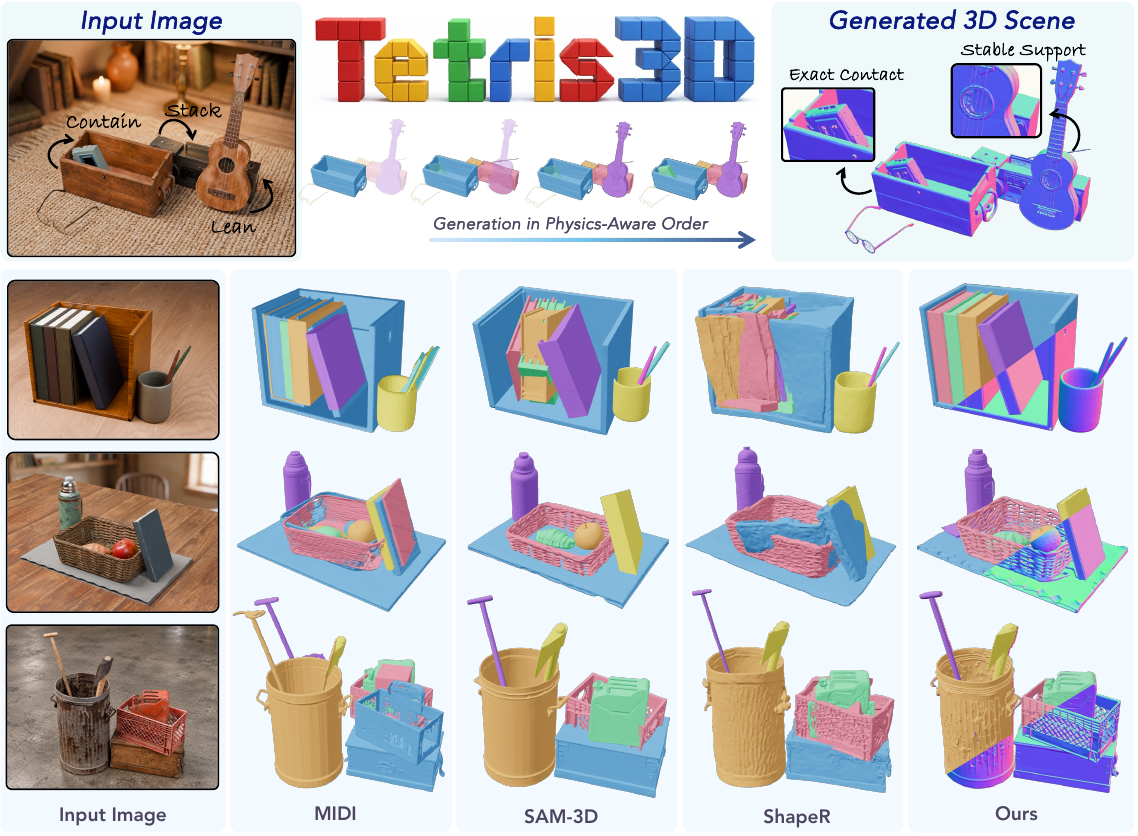}
    \caption{ \textbf{Teaser.}
    \textbf{\ours{}} reconstructs 3D scenes by generating objects that fit together as a scene. We generate objects autoregressively, explicitly conditioning each on neighboring geometry and physical relations to achieve geometric and physical coherence among objects, even under occlusion.}
\end{figure}

% from a image?

\begin{abstract}
% ver. final - 09/20
We propose \textbf{\ours{}}, a generative framework for single-image 3D scene reconstruction that recovers objects which are physically and geometrically coherent as a scene. Existing methods often generate objects independently or couple them implicitly, providing limited guidance for ensuring fine-grained spatial compatibility between neighboring objects that interact with one another. To address this, we explicitly condition the generation of each object on the geometry of surrounding objects and their physical relationships, guiding its shape and pose to remain geometrically and physically plausible within the scene. Moreover, we introduce \textit{ComOb}, a physics simulation-based dataset of 1.2M scenes featuring physical interactions across diverse object categories, with per-object meshes and pairwise physical relation annotations. Comprehensive experiments on synthetic and real-world scenes show that \ours{} recovers coherent object shapes and poses even when interacting regions are occluded, and achieves state-of-the-art performance in both generation quality and physical stability. 
% Our code, model weights, and dataset will be publicly released.

% We inject these conditions into a voxel-based generative backbone as spatially aligned, per-voxel signals, providing local guidance for generating geometry compatible with neighboring surfaces without requiring dedicated inter-object attention mechanisms. 

% We propose \textbf{\ours{}}, a generative framework for physically plausible 3D scene reconstruction from a single image. Existing methods often generate objects independently or couple them through implicit feature aggregation, providing limited guidance for the fine-grained geometric compatibility required by physical interactions. To address this, we explicitly condition each object generation on the geometry of neighboring objects and its physical relations to them. We inject these conditions into the voxel-based generative backbone as spatially aligned, per-voxel signals, providing local guidance for physically consistent generation without requiring dedicated inter-object attention mechanisms. Moreover, we introduce \textit{ComOb}, a physics simulation-based dataset of 1.2M scenes featuring physical interactions across diverse object categories, with per-object meshes and pairwise physical relation annotations. Comprehensive experiments on synthetic and real-world scenes show that \ours{} produces physically consistent reconstructions even when interacting regions are occluded and achieves state-of-the-art performance in both reconstruction quality and physical plausibility. Our code and model weights with dataset will be publicly released.

\end{abstract}

% \clearpage
% \input{sections/2_introduction}
% Ver. final - 09/20
\section{Introduction}
3D scene generation from an image reconstructs compositional scenes by recovering each object as a separate and complete 3D entity in a shared scene space. This enables the conversion of real-world observations into 3D assets for applications such as physical simulation, VR/AR, and robotic manipulation. To support these applications, scene reconstruction should recover not only plausible objects but also their physically and geometrically coherent composition. Interactions between objects---such as support and containment---depicted in the image require fine-grained spatial compatibility between neighboring objects. However, neighboring objects often occlude the very regions where they interact, making it challenging to recover this consistency from visible evidence alone.

Despite recent progress in 3D generation~\citep{zhang20233dshape2vecset, li2025triposg, zhang2024clay, xiang2026native, xiang2025structured}, recovering such inter-object spatial compatibility remains an open problem. Existing methods reconstruct scenes either by assembling individually generated objects~\citep{chen2026sam, zhou2026pose, li2026pixal3d, siddiqui2026shaper} or by jointly generating multiple objects or their poses through implicit feature aggregation~\citep{huang2025midi, meng2026scenegen, shi2026scenemaker}. These approaches provide limited explicit guidance for the local spatial compatibility required to recover the depicted interactions. 
% In either case, neighboring geometry and physical relationships are not used as explicit conditions for scene generation.

Motivated by this, we propose \textbf{\ours{}}, an object interaction-conditioned 3D generative framework for reconstructing scenes of objects that geometrically and physically fit together. Our key idea is to generate each object in relation to its neighbors rather than in isolation, by explicitly conditioning the generation of its shape and pose on neighboring geometry and physical relationships. Intuitively, when we interpret a scene, we use surrounding geometry to infer the space an object can occupy and physical relationships to reason how it should rest on or fit within its neighbors. Together, these cues help constrain possible object shapes and poses, even when the interacting regions are occluded. 
% The same neighbors that occlude an object can therefore also provide cues for reconstructing it.

% based on load-bearing support relations and derive the generation order by topologically sorting this hierarchy. 
Specifically, we reconstruct the scene by generating objects autoregressively, using previously generated neighbors as interaction context. We use a vision-language model (VLM) to infer a physical dependency hierarchy and derive the generation order by topologically sorting this hierarchy. For instance, given an image of a cup resting on a table, the table is generated first and its surface serves as the interaction context for the cup. For each object, we encode the neighboring geometry as \emph{spatial context}, represented by surface distance and vectors indicating relative position, and the physical relationships with neighboring objects as \emph{relation type}, represented by learnable embeddings. To inject these conditions in a manner spatially aligned with generation, we build on TRELLIS.2's voxel-based generative backbone~\citep{xiang2026native}. These interaction conditions are encoded on the same 3D grid used for generation and injected into the corresponding tokens, providing local guidance for generating each object shape that is spatially compatible with neighboring surfaces.

% For each object, we encode neighboring geometry as \emph{spatial context} using surface distances and relative-position vectors, and its physical relationships with neighboring objects as \emph{relation type} through learnable embeddings. To spatially align these conditions with the generation process, we build on the voxel-based generation of TRELLIS.2~\citep{xiang2026native}. These interaction conditions are encoded in the same 3D grid used for generation and injected into the corresponding generation tokens, providing local guidance for generating geometry compatible with neighboring surfaces.

Training \ours{} requires scenes that capture diverse object interactions with physical relation annotations. However, existing datasets are often restricted to specific environments~\citep{wang2026tableverse, ansari2026messykitchens}, or provide limited coverage of diverse physical interactions~\citep{fu20213d}. To address these limitations, we introduce \textit{ComOb}, a large-scale simulation-based scene dataset in which general objects are arranged according to diverse interaction types, such as stacking and containment, and settled into physically stable configurations through physics simulation~\citep{todorov2012mujoco}. It provides 1.2M scenes with per-object geometry, physical relation annotations, and occlusions that naturally arise from these arrangements. Finally, we conduct comprehensive experiments on synthetic~\citep{stojanov2021using} and real-world scenes~\citep{ansari2026messykitchens, yu2026picasso} to evaluate our framework. The results demonstrate that \ours{} reconstructs plausible scenes that preserve object interactions even when the interacting regions are occluded, achieving superior performance in reconstruction and generation quality, and physical stability.

\section{Preliminaries: TRELLIS.2} \label{Method:preliminaries}
% 9/8 V1 done
We build our framework on TRELLIS.2~\citep{xiang2026native}, which generates a 3D shape through a two-stage pipeline. It first generates the \emph{sparse structure} (SS), a binary occupancy grid \(\mathbf{O} \in \{0,1\}^{N_{s}\times N_{s}\times N_{s}}\) of resolution \(N_{s}\) in the canonical object space, whose \(L\) active voxels \(\mathcal{S}=\{\mathbf{p}_j\}_{j=1}^{L}\), with \(\mathbf{p}_j \in \{0,\dots,N_{s}-1\}^{3}\) the coordinate of the \(j\)-th active voxel, outline the coarse geometry. It subsequently generates the \emph{Structured LATents} (SLAT) \(\mathbf{z}^{\text{SLAT}}=\{\mathbf{z}_j\}_{j=1}^{L}\) attached to these active voxels, which encode geometry and texture. SLAT is the latent encoding of the O-Voxel representation, a field-free sparse voxel structure that encodes geometry and appearance jointly as per-voxel feature tuples \(\mathcal{F} = \{(\mathbf{f}^{\text{shape}}_{j}, \mathbf{f}^{\text{mat}}_{j}, \mathbf{p}_{j})\}_{j=1}^{L}\), where \(\mathbf{f}^{\text{shape}}_{j}\) and \(\mathbf{f}^{\text{mat}}_{j}\) describe the local geometry and material of the \(j\)-th voxel, respectively. Each stage is modeled by a diffusion transformer (DiT)~\citep{peebles2023scalable} with flow-matching formulation~\citep{lipman2022flow}. The first, \(v_\theta^{\text{SS}}\), denoises a dense latent volume on a coarse grid of resolution \(N_{c}\,(<N_{s})\), which is then decoded into \(\mathbf{O}\). The second, \(v_\theta^{\text{SLAT}}\), operates only on the \(L\) active voxels, treating them as tokens positionally embedded by their coordinates \(\mathbf{p}_j\). Both are conditioned on the input image features from a frozen DINOv3~\citep{simeoni2025dinov3} encoder. 
% Since TRELLIS.2 generates through an explicit 3D voxel grid as its intermediate representation, the space in which generation takes place is explicitly defined.
% We exploit this property: by embedding this grid in the scene space, we inject the conditions of object interaction and the back-projected image feature into the generation process in a spatially aligned, per-voxel manner.

\begin{figure}[t]
    \centering
    \includegraphics[width=\linewidth]{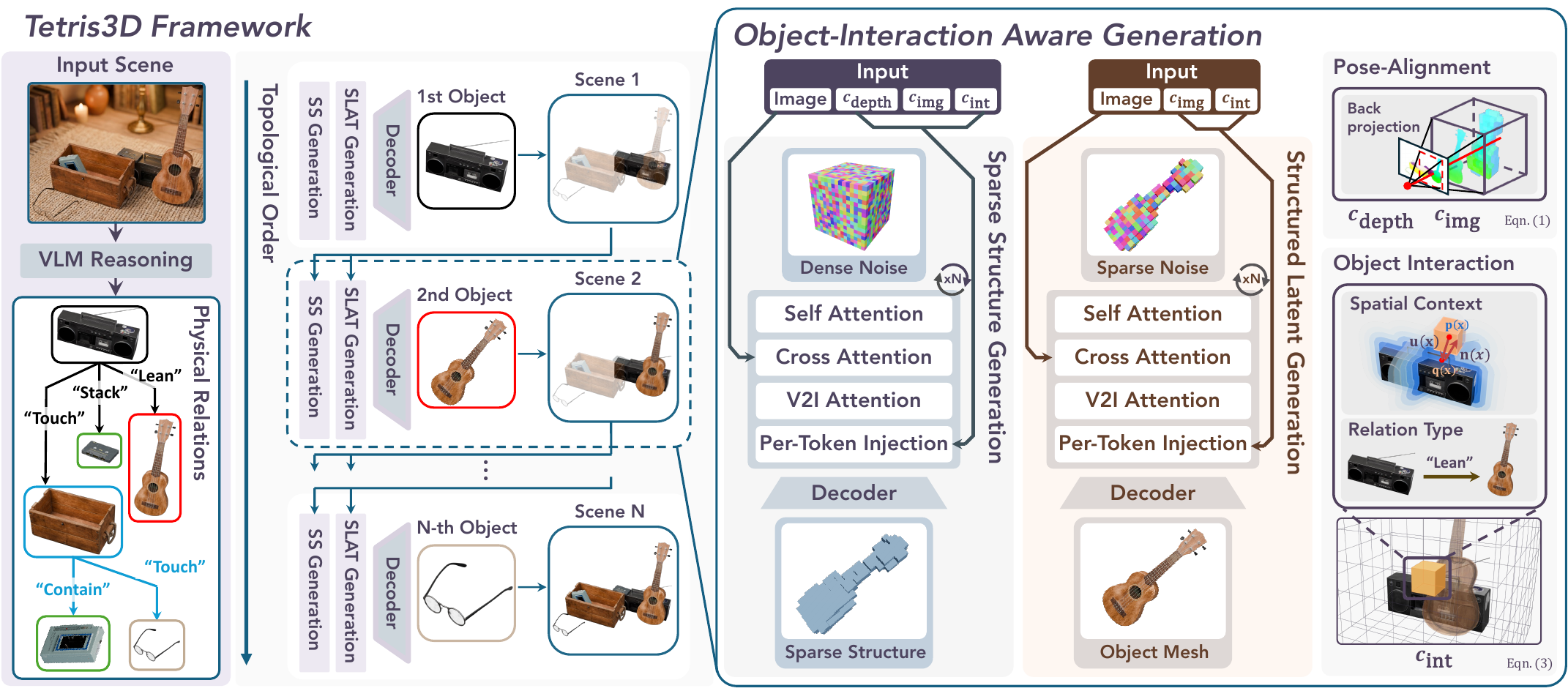}
    \caption{
        % \ours{}는 VLM으로부터 추론된 physical relation과 object들간의 dependency를 바탕으로 구성된 topolgical order로 scene을 autoregressive하게 생성한다. 각 object generation은 follows a two-stage generation pipeline consisting of sparse structure generation and structured latent generation, with two novel components in each DiT: (1) \emph{Per-Token Injection}, we derive \(c_{\text{depth}}\) from depth map, \(c_{\text{depth}}\) from back-projected image features, and \(c_{\text{int}}\) from object-interaction on the same 3D grid used for generation. These are injected into the corresponding tokens for pose-aligned, interaction-aware generation. (2) \emph{V2I Attention}, the target object's visible mask partitions the 3D grid into visible and invisible tokens, and cross-attention flows from visible to invisible tokens.
        \textbf{Overview of \ours{}.} \ours{} reconstructs scenes autoregressively in a topological order determined by VLM-inferred physical relations and dependencies (Sec.~\ref{Method:training-inference}). Each object is generated through a two-stage pipeline consisting of sparse structure and structured latent generation, with two key components in each DiT: (1) \emph{Per-Token Injection} (Sec.~\ref{Method:scene-conditioning}): We derive \(c_{\text{depth}}\) from the depth map, \(c_{\text{img}}\) from back-projected image features, and \(c_{\text{int}}\) from object interactions on the same 3D grid used for generation and these conditions are injected into the corresponding tokens. (2) \emph{V2I Attention} (Sec.~\ref{Method:v2iv-attn}): Invisible tokens identified by object mask attend to visible tokens through cross-attention, using the aggregated information to complete occluded regions.}
    \vspace{-0.3cm}
    \label{fig:main_architecture}
\end{figure}

 % to guide pose-aligned, interaction-aware generation
% The target object's visible mask identifies visible and invisible tokens through camera projection. 

\section{Method}

\subsection{Overview}
\label{Method:overview}
% ver. final - 09/20
We propose \textbf{\ours{}}, a generative framework for object-centric 3D scene reconstruction from a single image. Our goal is to reconstruct a scene in which objects are physically and geometrically coherent with one another by explicitly conditioning generation on surrounding geometry and physical relationships. To achieve this, we incorporate object interactions into the generation process in two forms: \emph{spatial context} and \emph{relation type}. The \emph{spatial context} captures the geometric arrangement of neighboring objects relative to the target object, helping recover geometry spatially compatible with the surroundings and avoid physically invalid configurations such as penetration. The \emph{relation type} specifies the physical relationship between the target object and its neighbors, guiding the generated shape to reflect the interactions depicted in the input scene. Furthermore, to reconstruct the scene without a separate per-object pose-estimation stage, we generate each object directly in the pose observed in the scene image using the depth map and image feature back-projection. We also introduce a \emph{visible-to-invisible (V2I) cross-attention} that guides the completion of occluded regions using information aggregated from visible tokens. To generate a scene, we generate objects autoregressively in a physical dependency order inferred by a VLM, so that each object is generated after the neighbors it depends on. We describe the overall pipeline in Fig.~\ref{fig:main_architecture}.

% We propose \textbf{\ours{}}, a generative framework for object-centric 3D scene reconstruction from a single image. Our goal is to reconstruct a physically consistent scene by explicitly conditioning generation on spatial and physical interactions among the objects. To achieve this, we incorporate physical interactions into the generation process in two aspects: \emph{spatial context} and \emph{relation type}. The \emph{spatial context} captures the geometric arrangement of neighboring objects relative to the target object, helping avoid physically invalid configurations such as penetration. The \emph{relation type} specifies the physical relationship between the target object and its neighbors, guiding the generated shape to reflect the interactions depicted in the input scene. Furthermore, to reconstruct the scene without a separate per-object pose-estimation stage, we generate each object directly in the pose observed in the scene image using the depth map and image feature back-projection. We also introduce a visible-to-invisible cross-attention mechanism that guides the completion of occluded regions using information aggregated from visible tokens. We describe the overall pipeline in Fig.~\ref{fig:main_architecture}.

% we first briefly review TRELLIS.2~\citep{xiang2026native}, which serves as a backbone of our framework (Sec.~\ref{Method:preliminaries}). 
% In the following, we first describe pose-aligned object generation in the scene and the object-interaction conditioning mechanism (Sec.~\ref{Method:scene-conditioning}). 
In the following, we first describe pose-aligned and interaction-aware object generation in the scene (Sec.~\ref{Method:scene-conditioning}).
Next, we present \emph{visible-to-invisible cross-attention} introduced for faithful completion of occluded parts (Sec.~\ref{Method:v2iv-attn}). Finally, we present \textit{ComOb}, a physically consistent simulation-based scene dataset (Sec.~\ref{Method:comob}), and describe the inference pipeline (Sec.~\ref{Method:training-inference}).

\subsection{\ours{}: Geometrically and Physically Coherent 3D Scene Generation} \label{Method:scene-conditioning}
\begin{wrapfigure}{hr}{0.4\textwidth}
    \centering
    \vspace{-11pt}
    \includegraphics[width=\linewidth]{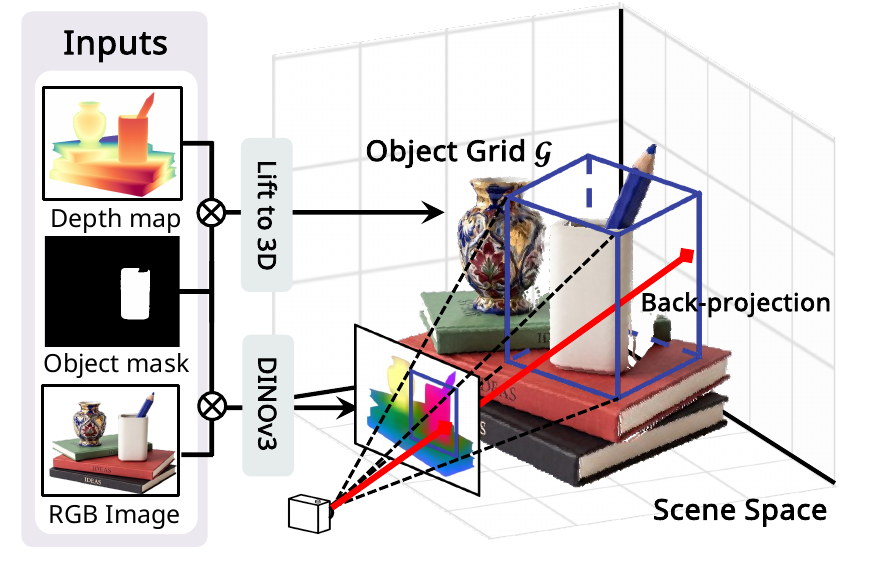}
    \vspace{-8pt}
    \caption{\textbf{Schematic of object grid \(\mathcal{G}\) and pose-aligned generation.}}
    \label{figure:subfigure_camera_geometry}
    \vspace{-10pt}
\end{wrapfigure}

% 09/08 V3. w/ JM version
\subsubsection{Pose-Aligned Object Generation in the Scene} \label{Method:scene-camera-ray}
% 9/9 version 3
Unlike previous methods that separately estimate poses of generated shapes~\citep{shi2026scenemaker, chen2026sam} or require post-alignment~\citep{li2026pixal3d, zhou2026pose}, we directly generate shapes in scene space with poses aligned with the input scene image. Here, scene space refers to the input camera's 3D coordinate system, shared by all objects in the image. As shown in Fig.~\ref{figure:subfigure_camera_geometry}, we first define an object grid from the depth map (sensor-captured or estimated), where the target object is generated. Inspired by Pixal3D~\citep{li2026pixal3d}, we back-project image features into this grid along camera rays to guide shape generation in alignment with the input scene image. Specifically, given the scene image, target object mask \(M\), and depth map, we lift the pixels within \(M\) into a 3D partial point cloud \(P\) using their depth. We then define the object grid \(\mathcal{G}\) over a cube that encloses \(P\) with an additional margin. To guide the position and orientation of the object in this grid, we construct a voxel-wise condition from the partial point cloud \(P\) and the DINOv3~\citep{simeoni2025dinov3} feature map \(F\). Let \(\mathcal{G}_P \subseteq \mathcal{G}\) denote the set of voxels containing at least one point from \(P\).
For each voxel \(x \in \mathcal{G}\), we define:
\begin{equation} \label{equ:pose-alignment}
\begin{aligned}
c_{\mathrm{depth}}(x)
&=
\begin{cases}
\mathbf{e}_{\mathrm{depth}}, & x \in \mathcal{G}_P, \\
\mathbf{0}, & \text{otherwise},
\end{cases} \qquad c_{\mathrm{img}}(x)
&= F\big(\pi(x)\big),
\end{aligned}
\end{equation}
where \(\mathbf{e}_{\mathrm{depth}}\) is a learnable embedding indicating observed surface occupancy, and \(\pi\) projects the voxel center from scene space to feature map coordinates along the camera ray. Together, these signals determine the object pose in the grid, so the generated shape is aligned with the scene space.

% \vspace{-1em}
\subsubsection{Object Interaction-Aware Generation} \label{Method:object-interaction}
% 0909 v3 based on JM feedback
We incorporate two types of information into an object-interaction condition: (1) \emph{spatial context}, which discourages interpenetration and floating for physical plausibility, and (2) \emph{relation type}, which guides the generated shape to reflect the physical interactions depicted in the input scene image. 
\paragraph{Spatial Context.} To accommodate non-watertight meshes, we represent spatial context using an unsigned distance field (UDF), complemented with the direction to the nearest surface point and the surface normal at that point. Together, these quantities describe the voxel's spatial relationship to the neighboring surfaces. Specifically, given a target object grid \(\mathcal{G}\) and a set of neighboring objects \(\mathcal{H}\), let \(\mathcal{H}_{\mathrm{union}}\) denote the union of the surfaces of the objects in \(\mathcal{H}\). For each voxel \(x \in \mathcal{G}\) with its center coordinate \(\mathbf{p}(x) \in \mathbb{R}^3\), let \(\mathbf{q}(x) \in \mathcal{H}_{\mathrm{union}}\) be its nearest surface point. We define the truncated unsigned distance \(d(x)\), the direction vector \(\mathbf{u}(x)\) from the voxel center coordinate to the nearest surface point, and the surface normal of the nearest surface point \(\mathbf{n}(x)\) as:
\begin{equation}
d(x) = \min\big(\|\mathbf{p}(x)-\mathbf{q}(x)\|,\,\tau\big), \quad
\mathbf{u}(x) = -\frac{\mathbf{p}(x)-\mathbf{q}(x)}
{\|\mathbf{p}(x)-\mathbf{q}(x)\|}, \quad 
\mathbf{n}(x) = \mathbf{n}_{\mathcal{H}_{\mathrm{union}}}\big(\mathbf{q}(x)\big),
\end{equation}
where \(\tau\) is the truncation distance.
\paragraph{Relation Type.} We encode the physical relation between the target object and each neighboring object using a learnable embedding from a predefined set of relation types, \(\mathcal{R}=\{\textit{stack},\textit{lean},\textit{contain},\textit{touch},\textit{none}\}\). The relation types are either provided as input or inferred by a VLM together with the physical dependency reasoning described in Sec.~\ref{Method:training-inference}. For each voxel \(x\), let \(r(x) \in \mathcal{R}\) denote the relation between the target object and the neighboring object to which the nearest surface point \(\mathbf{q}(x)\) belongs. 
\paragraph{Interaction Condition.} We combine the spatial context and relation type into the per-voxel interaction condition as:
\begin{equation}
c_{\mathrm{int}}(x) =
\begin{cases}
\operatorname{MLP}\Big(\Big[\big(1-\tfrac{d(x)}{\tau}\big)\,\mathbf{e}_{\mathrm{dist}},\;\;
\operatorname{MLP}\big([\mathbf{u}(x);\,\mathbf{n}(x)]\big)\Big]\Big) + \mathbf{W}_{\mathrm{rel}}\big[r(x)\big] & \text{if } d(x) < \tau, \\[4pt]
\mathbf{0} & \text{otherwise},
\end{cases}
\end{equation}
where \(\mathbf{e}_{\mathrm{dist}}\) is a learnable distance embedding, and \(\mathbf{W}_{\mathrm{rel}}\) is a learnable relation embedding table. The proximity factor \( (1-d(x)/\tau)\) assigns a larger weight to the distance embedding for voxels closer to a neighboring surface. Voxels at or beyond the truncation distance \(\tau\) receive no condition.

\subsubsection{Voxel-Aligned Condition Injection} \label{Method:architecture} \label{Method:per-token-injection}
% 0909 v3 JM feedback
Since \(c_{\mathrm{img}}, c_{\mathrm{depth}}\) and \(c_{\mathrm{int}}\) are defined on the same object grid \(\mathcal{G}\), all generation and conditioning tokens are spatially aligned by construction. By exploiting this alignment, we inject those conditions directly into the corresponding tokens at every DiT layer of \(v_\theta^{\text{SS}}\) and \(v_\theta^{\text{SLAT}}\). Specifically, at the \(l\)-th layer, these conditions are each projected by an MLP and added to the token feature:
\begin{equation}
\mathbf{h}^{(l)}(x) \leftarrow \mathbf{h}^{(l)}(x) + \operatorname{MLP}^{(l)}_{\mathrm{img}}\big(c_{\mathrm{img}}(x)\big) + \lambda_{\mathrm{depth}} \cdot \operatorname{MLP}^{(l)}_{\mathrm{depth}}\big(c_{\mathrm{depth}}(x)\big) + \lambda_{\text{int}} \cdot \operatorname{MLP}^{(l)}_{\mathrm{int}}\big(c_{\mathrm{int}}(x)\big),
\end{equation}
where \(\mathbf{h}^{(l)}(x)\) denotes the feature of the token at voxel \(x\), and \(\lambda_{\mathrm{depth}}\), \(\lambda_{\mathrm{int}}\) are learnable scalar gates initialized to zero. This initialization leaves the original image-conditioned update unchanged at the start of finetuning, while allowing the model to adaptively incorporate the additional condition during training~\citep{alayrac2022flamingo}. 
This update is applied to all tokens of the dense latent volume in sparse structure generation, and only to tokens at active voxels in structured latent generation.
%For sparse structure generation, this update is applied to all tokens in the dense latent volume and only to tokens at active voxels for structured latent generation.

\subsection{V2I Attention: Amodal Completion via Visible-to-Invisible Attention} ~\label{Method:v2iv-attn}
The main challenge in amodal generation from an occluded image is that the invisible region lacks conditional guidance. To mitigate this, inspired by completion and inpainting tasks~\citep{li2023proxyformer, wei2023diffusion}, we introduce \emph{visible-to-invisible (V2I) cross-attention}, an explicit mechanism that updates the occluded part from the visible part. Specifically, since the tokens in the invisible region can be inferred from the tokens in the visible region, we adopt an additional cross-attention layer with the invisible tokens as queries and the visible tokens as keys and values. For the object grid \(\mathcal{G}\) described in Sec.~\ref{Method:scene-camera-ray}, let \(\mathcal{T}\) denote a set of tokens corresponding to each voxel in \(\mathcal{G}\): all dense grid tokens for sparse structure generation, and the active-voxel tokens for structured latent generation. For a target object \(o_i\), we define the occlusion mask as \(M_i^{\mathrm{occ}}=\bigcup_{j\neq i} M_j\), treating all pixels occupied by other objects as potential occlusions of the target. The tokens whose voxels lie along the camera rays through these pixels form the invisible token set \(\mathcal{T}_{\mathrm{invis}}\), and all remaining tokens form the visible token set \(\mathcal{T}_{\mathrm{vis}} = \mathcal{T} \setminus \mathcal{T}_{\mathrm{invis}}\). At every DiT layer \(l\) of both \(v_{\theta}^{\mathrm{SS}}\) and \(v_{\theta}^{\mathrm{SLAT}}\), we update the invisible tokens by cross-attending to the visible tokens:
\begin{equation}
\mathbf{h}^{(l)}_{\mathcal{T}_{\mathrm{invis}}}
\leftarrow
\mathbf{h}^{(l)}_{\mathcal{T}_{\mathrm{invis}}}
+
\lambda^{(l)}_{\mathrm{V2I}}
\cdot
\operatorname{CrossAttn}
\big(
\mathbf{h}^{(l)}_{\mathcal{T}_{\mathrm{invis}}},\;
\mathbf{h}^{(l)}_{\mathcal{T}_{\mathrm{vis}}}
\big),
\end{equation}
where \(\lambda^{(l)}_{\mathrm{V2I}}\) is a learnable scalar initialized to zero.

\subsection{ComOb: Simulation-Based Composited Object Scene Dataset} \label{Method:comob}
We construct a large-scale simulation-based dataset in which diverse objects engage in physical interactions. Existing datasets are often restricted to specific environments, such as indoor scenes~\citep{fu20213d} or tabletops~\citep{wang2026tableverse, ansari2026messykitchens}, and provide limited coverage of complex physical interactions~\citep{fu20213d}. These limitations make existing datasets less suitable for learning physical interactions among general objects. To address this, we introduce \textit{ComOb}, a simulation-based physically consistent dataset of composite object scenes that captures physical interactions across diverse object categories, together with interaction-induced occlusions. To ensure that each scene contains direct physical interactions, we construct scenes around a predefined set of interaction types: \(\{\textit{stack}, \textit{lean}, \textit{contain}, \textit{pile}, \textit{touch}\}\). In each scene, target objects are arranged with respect to 3D primitives, such as cylinders and cones, according to the given interaction type, and are then settled into physically stable configurations through a physics engine~\citep{todorov2012mujoco}. Each target--primitive pair is annotated with its corresponding physical relation \(r \in \mathcal{R}\), as defined in Sec.~\ref{Method:object-interaction}. We use target objects from TRELLIS-500K~\citep{xiang2025structured} and, in total, generate 1.2M scenes and 3.4M annotated samples from the generated scenes, comprising rendered RGB images, object masks, depth maps, object meshes, and physical relation labels. Further details are provided in Appendix~\ref{Appendix:dataset-details}.

\subsection{Inference Pipeline} \label{Method:training-inference}
Our framework reconstructs a scene by sequentially generating objects conditioned on their interactions with other objects in the same scene. At inference, when reconstructing a scene from scratch, no reconstructed object is initially available for interaction conditioning. Therefore, we generate objects autoregressively in an order that respects their \emph{physical dependencies}. Specifically, following prior works~\citep{li2026phi, chen2026physcene3d, lin2026pat3d}, we use a VLM~\citep{bai2025qwen3} to infer a directed physical dependency hierarchy, and derive the generation order by topologically sorting this hierarchy. Each node represents an object, and a directed edge from object \(k\) to object \(i\) indicates that \(i\) relies on \(k\) for physical support or stability and should therefore be generated after \(k\). Each edge is additionally labeled with the corresponding physical relation \(r \in \mathcal{R}\). For example, a cup resting on a table depends on the table for support against gravity. We therefore generate the table first and use its reconstructed surface as physical context when generating the cup. By generating supporting objects before their dependents, the geometry required for interaction conditioning becomes available progressively throughout the autoregressive reconstruction. We provide further details in Appendix~\ref{Appendix:inference-details}.

\begin{table*}[t]
    \centering
    \caption{
        \textbf{Quantitative comparison on Toys4K~\citep{stojanov2021using}}. \(\dagger\) denotes methods whose generated objects are aligned to the scene using FoundationPose~\citep{wen2024foundationpose}.
        }
    \label{tab:toys4k_main_comparison}

    \begingroup
    \fontsize{6}{7}\selectfont
    \setlength{\tabcolsep}{2pt}
    \renewcommand{\arraystretch}{1.15}

    \resizebox{\textwidth}{!}{%
    % Method (1) + Reconstruction (6) + Generation (5)
    % + Physical Plausibility (3) = 15 columns
    \begin{tabular}{@{}l*{14}{c}@{}}
        \toprule
        %
        % Header 1: Metric groups
        \multirow{2}{*}{\textbf{Method}}
        & \multicolumn{6}{c}{\textbf{Reconstruction Quality}}
        & \multicolumn{5}{c}{\textbf{Generation Quality}}
        & \multicolumn{3}{c}{\textbf{Physical Stability}} \\
        \cmidrule(lr){2-7}
        \cmidrule(lr){8-12}
        \cmidrule(lr){13-15}
        %
        % Header 2: Individual metrics
        % Reconstruction Quality
        & CD-S~\(\downarrow\)
        & CD-O~\(\downarrow\)
        & F1-S~\(\uparrow\)
        & F1-O~\(\uparrow\)
        & IoU-B~\(\uparrow\)
        & ICP-Rot~\(\downarrow\)
        % Generation Quality
        & MMD~\(\downarrow\)
        & COV~\(\uparrow\)
        & P-FID~\(\downarrow\)
        & Uni3D~\(\uparrow\)
        & ULIP~\(\uparrow\)
        % Physical Plausibility
        & PD~\(\downarrow\)
        & \(D_{\mathrm{mean}}\,\downarrow\)
        & \(E_{\mathrm{peak}}\,\downarrow\) \\
        \midrule
        \multicolumn{15}{c}{
            \textbf{\textit{Scene Generation}}
        } \\
        \midrule
        SAM-3D
        & 25.19 & 1.70 & 0.2462 & 0.6296 & 0.3817 & 20.37 & 2.840 & 71.24 & 5.222 & \underline{0.6004} & \underline{0.6314}
        & 1.354 & 192.4 & 0.8633 \\             % Physical
        %
        % PixARMesh
        % & -- & -- & -- & -- & -- & --
        % & -- & -- & -- & -- & --
        % & -- & -- & -- \\
        %
        ShapeR
        & \underline{5.72} & \underline{1.61} & \underline{0.6480} & \underline{0.6302} & \underline{0.6408} & \underline{10.65} & \underline{2.590} & \underline{71.61} & 12.45 & 0.5165 & 0.5594
        & 0.8862 & 115.1 & \underline{0.4675} \\
        WorldSculpt
        & 8.68 & 1.98 & 0.4302 & 0.5976 & 0.5369 & 17.31 & 2.973 & 70.56 & \underline{4.816} & 0.5454 & 0.5776
        & \underline{0.1177} & \underline{103.5} & 0.4812 \\
        \midrule
        \multicolumn{15}{c}{
            \textbf{\textit{Amodal Generation + Pose Estimation}}%
        } \\
        \midrule
        Amodal3R\( ^{\dagger}\)
        & 12.51 & 2.25 & 0.4455 & 0.5546 & 0.4634 & 22.97 & 3.331 & 65.66 & 6.998 & 0.5390 & 0.5720
        & 0.5962 & 115.0 & 0.5457 \\
        GENA3D\( ^{\dagger}\)
        & 15.91 & 2.73 & 0.3812 & 0.5091 & 0.4126 & 27.63 & 3.928 & 62.46 & 8.280 & 0.4743 & 0.5015
        & 0.9106 & 131.0 & 0.6193 \\
        \midrule
        \textbf{Ours}
        & \textbf{3.68} & \textbf{0.97} & \textbf{0.8407} & \textbf{0.8084} & \textbf{0.7918} & \textbf{5.29}
        & \textbf{1.501} & \textbf{83.11} & \textbf{1.599} & \textbf{0.6884} & \textbf{0.7182}
        & \textbf{0.0036} & \textbf{38.91} & \textbf{0.1721} \\
        \bottomrule
    \end{tabular}%
    }
    \endgroup
\end{table*}

\begin{table*}[t]
    \centering
    \caption{
        \textbf{Quantitative comparison on MessyKitchens~\citep{ansari2026messykitchens} and Picasso~\citep{yu2026picasso}.} \(\dagger\) denotes methods whose generated objects are aligned to the scene using FoundationPose~\citep{wen2024foundationpose}.
    }
    \label{tab:scene_comparison}

    \begingroup
    \tiny
    \setlength{\tabcolsep}{1.2pt}
    \renewcommand{\arraystretch}{1.1}

    \resizebox{\textwidth}{!}{%
    % Method (1) + MessyKitchen (9) + Benchmark2 (9) = 19 columns
    \begin{tabular}{@{}l*{18}{c}@{}}
        \toprule

        % Dataset
        \multirow{3}{*}{\textbf{Method}}
        & \multicolumn{9}{c}{\textbf{MessyKitchens}}
        & \multicolumn{9}{c}{\textbf{Picasso
        }} \\

        \cmidrule(lr){2-10}
        \cmidrule(lr){11-19}

        % Metric group
        & \multicolumn{6}{c}{\textbf{Reconstruction Quality}}
        & \multicolumn{3}{c}{\textbf{Physical Stability}}
        & \multicolumn{6}{c}{\textbf{Reconstruction Quality}}
        & \multicolumn{3}{c}{\textbf{Physical Stability}} \\

        \cmidrule(lr){2-7}
        \cmidrule(lr){8-10}
        \cmidrule(lr){11-16}
        \cmidrule(lr){17-19}

        % Metrics
        & CD-S~\(\downarrow\)
        & CD-O~\(\downarrow\)
        & F1-S~\(\uparrow\)
        & F1-O~\(\uparrow\)
        & IoU-B~\(\uparrow\)
        & ICP-Rot~\(\downarrow\)
        & PD~\(\downarrow\)
        & \(D_{\mathrm{mean}}\downarrow\)
        & \(E_{\mathrm{peak}}\downarrow\)

        & CD-S~\(\downarrow\)
        & CD-O~\(\downarrow\)
        & F1-S~\(\uparrow\)
        & F1-O~\(\uparrow\)
        & IoU-B~\(\uparrow\)
        & ICP-Rot~\(\downarrow\)
        & PD~\(\downarrow\)
        & \(D_{\mathrm{mean}}\downarrow\)
        & \(E_{\mathrm{peak}}\downarrow\) \\

        \midrule
        \multicolumn{19}{c}{\textbf{\textit{Scene Generation}}} \\
        \midrule

        SAM-3D
        & 0.25 & \underline{0.64} & 0.8749 & \underline{0.8187} & 0.5417 & 9.43 & 0.0270 & \underline{215.9} & \underline{0.9949}
        & 0.66 & \underline{1.12} & 0.7432 & \underline{0.7254} & 0.5631 & 12.87 & 0.1221 & 1033.4 & 6.389 \\

        % PixARMesh
        % & -- & -- & -- & -- & -- & -- & -- & -- & --
        % & -- & -- & -- & -- & -- & -- & -- & -- & -- \\

        ShapeR
        & \underline{0.23} & 1.57 & \underline{0.9089} & 0.6976 & \underline{0.6152} & \underline{9.38} & 0.0253 & 388.8 & 1.825
        & 0.65 & 1.90 & \underline{0.7880} & 0.6293 & \underline{0.6887} & \underline{12.04} & 0.0666 & 932.2 & 4.335 \\

        WorldSculpt
        & 0.32 & 1.19 & 0.8601 & 0.7445 & 0.5552 & 9.88 & \underline{0.0012} & 266.3 & 1.291
        & \underline{0.54} & 1.69 & 0.7834 & 0.6975 & 0.5087 & 18.66 & \underline{0.0111} & 900.3 & 4.125 \\

        MIDI
        & 2.79 & 2.85 & 0.3966 & 0.5725 & 0.1361 & 36.10 & 1.606 & 488.4 & 2.669 
        & 1.60 & 2.47 & 0.6063 & 0.5398 & 0.3331 & 32.23 & 8.058 & 887.5 & 4.972\\

        SceneGen
        & 2.47 & 2.99 & 0.4232 & 0.5481 & 0.0682 & 25.70 & 27.21 & 2354.5 & 11.40 & 2.54 & 3.18 & 0.4509 & 0.4841 & 0.2132 & 40.04 & 30.83 & 2784.5 & 12.52 \\

        SceneMaker
        & 1.35 & 2.00 & 0.5394 & 0.6113 & 0.2858 & 25.48 & 2.069 & 531.8 & 2.527 & 1.78 & 2.24 & 0.5395 & 0.5611 & 0.4222 & 24.62 & 4.636 & 915.3 & 4.488 \\

        \midrule
        \multicolumn{19}{c}{
            \textbf{\textit{Amodal Generation + Pose Estimation}}
        } \\
        \midrule

        Amodal3R\( ^{\dagger}\)
        & 0.32 & 1.66 & 0.8787 & 0.6996 & 0.5695 & 9.52 & 0.0447 & 267.5 & 1.407 & 0.66 & 1.75 & 0.7826 & 0.7121 & 0.6203 & 17.01 & 0.1397 & \underline{875.5} & \underline{4.060} \\

        GENA3D\( ^{\dagger}\)
        & 0.46 & 2.08 & 0.8253 & 0.6636 & 0.5315 & 11.61 & 0.0265 & 290.1 & 1.668 & 1.27 & 2.61 & 0.6494 & 0.5233 & 0.4384 & 32.87 & 0.1416 & 1006.6 & 5.083 \\

        \midrule

        \textbf{Ours}
        & \textbf{0.11} & \textbf{0.63} & \textbf{0.9620} & \textbf{0.8284} & \textbf{0.7113} & \textbf{5.19} & \textbf{0.0011} & \textbf{102.0} & \textbf{0.3347}
        & \textbf{0.27} & \textbf{1.06} & \textbf{0.9053} & \textbf{0.7325} & \textbf{0.7264} & \textbf{8.913} & \textbf{0.0016} & \textbf{434.4} & \textbf{2.147} \\

        \bottomrule
    \end{tabular}%
    }

    \endgroup
\end{table*}

\section{Experiments}
We describe implementation details in Appendix~\ref{Appendix:implementation-detail} and evaluation details in Appendix~\ref{Appendix:evaluation-detail}. We further provide additional analyses in Appendix~\ref{Appendix:additional-results}, including results with estimated depth and VLM-based relation reasoning. Finally, we compare our method with agentic generation using GPT-6 Astra~\citep{openai2026gpt6astra} in Appendix~\ref{Appendix:comparison-astra}.

% including the effect of estimated depth and VLM-based relation reasoning are provided in Appendix~\ref{Appendix:additional-results}. 

% We further provide an additional comparison on GSO~\citep{downs2022google}, with additional analyses including comparison on occluded region, the effect of estimated depth and VLM-based relation reasoning are provided in Appendix~\ref{Appendix:additional-results}. 

\paragraph{Baselines.}
% We compare \ours{} against scene generation baselines, including SAM-3D~\citep{chen2026sam}, ShapeR~\citep{siddiqui2026shaper}, WorldSculpt~\citep{niu2026worldsculpt}, SceneMaker~\citep{shi2026scenemaker}, SceneGen~\citep{meng2026scenegen}, and MIDI~\citep{huang2025midi}. We further compare against amodal object generation methods, Amodal3R~\citep{wu2025amodal3r} and GENA3D~\citep{zhou2025gena3d}, each paired with a FoundationPose~\citep{wen2024foundationpose} to place the objects in the scene.
We compare \ours{} against scene generation methods, including SAM-3D~\citep{chen2026sam}, ShapeR~\citep{siddiqui2026shaper}, WorldSculpt~\citep{niu2026worldsculpt}, SceneMaker~\citep{shi2026scenemaker}, SceneGen~\citep{meng2026scenegen}, and MIDI~\citep{huang2025midi}. We compare against amodal generation methods, Amodal3R~\citep{wu2025amodal3r} and GENA3D~\citep{zhou2025gena3d}, paired with FoundationPose~\citep{wen2024foundationpose} for placement in the scene.

% We evaluate \ours{} on 3 benchmarks: Toys4K~\citep{stojanov2021using} and GSO~\citep{downs2022google}, whose object sources are disjoint from training, and MessyKitchens~\citep{ansari2026messykitchens} which consists of real and contact-rich scenes. To evaluate in scenarios involving interaction and occlusion, 우리는 ComOb generator를 활용하여 per-object로 scene을 만들었다. We assess the generated scenes along 3 perspectives: (i) \emph{scene-level quality}, how well the objects are jointly reconstructed in the scene space in terms of both shape and pose, and (ii) \emph{object-level quality}, how accurately the geometry of each individual object is recovered in its canonical space, and (iii) \emph{generative quality}, how realistic and semantically faithful the generated shapes are, which is measured at the level of distributions and learned representations rather than by point-wise comparison against a matched object.

\paragraph{Evaluation Protocol.}

We evaluate \ours{} on Toys4K~\citep{stojanov2021using}, whose object sources are disjoint from those used for training.
% To evaluate reconstruction under object interaction and occlusion, we compose evaluation scenes from Toys4K using our \textit{ComOb} generator.  
To evaluate reconstruction under object interaction and occlusion, we compose evaluation scenes by arranging primitives around each Toys4K object with our \textit{ComOb} generator.
On these scenes, all methods receive the ground-truth depth maps; results with depth estimated by MoGe3~\citep{kong2026moge} are reported in Appendix~\ref{Appendix:additional-results}. We further compare \ours{} against baselines on Picasso~\citep{yu2026picasso} and MessyKitchens~\citep{ansari2026messykitchens}, which contain real and contact-rich scenes, to assess generalization beyond \textit{ComOb}-generated settings. We use estimated depth from MoGe3 and physical relations inferred by a VLM~\citep{bai2025qwen3} and follow our inference pipeline described in Sec.~\ref{Method:training-inference}. We evaluate the reconstructed scenes from three perspectives: (i) \emph{scene-level quality}, measuring how accurately object shapes and poses are jointly recovered in scene space; (ii) \emph{object-level quality}, measuring the geometric fidelity of each object in the object canonical space; and (iii) \emph{physical stability}, measuring the physical consistency of the reconstructed scene through inter-object penetration and post-simulation stability. The \emph{simulation stability} evaluates whether objects remain in place without falling or drifting when the reconstructed scene is simulated in a physics engine~\citep{todorov2012mujoco}.

% We evaluate \ours{} on Toys4K~\citep{stojanov2021using} and GSO~\citep{downs2022google}, whose object sources are disjoint from training, To evaluate in scenarios involving interaction and occlusion, we compose the evaluation scenes from the objects of each benchmark using our ComOb generator. We also compare \ours active baselines on Picasso~\citep{yu2026picasso} and MessyKitchens~\citep{ansari2026messykitchens} to evaluate the generalization?, which consists of real, contact-rich scenes. We assess the generated scenes along 3 perspectives: (i) \emph{scene-level quality}, how well the objects are jointly reconstructed in the scene space in terms of both shape and pose, (ii) \emph{object-level quality}, how accurately the geometry of each individual object is recovered in its canonical space, and (iii) emph{physical stability}, how physically consistent the generated scene is, measured by penetration depth and by post-simulation stability that measures the objects remain in place without falling or drifting when the scene is loaded into a physics simulation engine~\citep{todorov2012mujoco}.

\paragraph{Metrics.}
We measure \emph{scene-level quality} using scene-level Chamfer Distance (CD-S) and F-Score (F1-S), bounding-box IoU (IoU-B), and the residual rotation error after per-object ICP alignment (ICP-Rot), and \emph{object-level quality} using object-level Chamfer Distance (CD-O) and F-Score (F1-O). CD is reported \( \times 10^2\). For \emph{generative quality}, we report MMD and COV~\citep{achlioptas2018learning} against the ground-truth object distribution, P-FID~\citep{nichol2022point}, the Fr\'echet distance between point-cloud features of generated and ground-truth objects, and ULIP~\citep{xue2024ulip} and Uni3D~\citep{zhou2024uni3d} similarities between the input image and generated point cloud as semantic measures. MMD is reported in units of \(10^{-3}\), while COV is reported as a percentage. For \emph{physical stability}, following \citep{lee2026simuscene, li2026phi}, we report penetration depth (PD) to quantify interpenetration, mean displacement (\(D_{\text{mean}}\)) to measure settling error, and peak kinetic energy per unit mass (\(E_{\text{peak}}\), in \(\mathrm{J/kg}\)) to measure dynamic instability during simulation.
% For \emph{physical stability}, we report penetration depth (Pen. Depth) for , mean displacement \(D_{\text{mean}}\) for settling errors, Peak Energy (\(E_{\text{peak}}\)) for dynamic artifacts.

\begin{figure*}[!t]
    \centering
    \includegraphics[width=\linewidth]{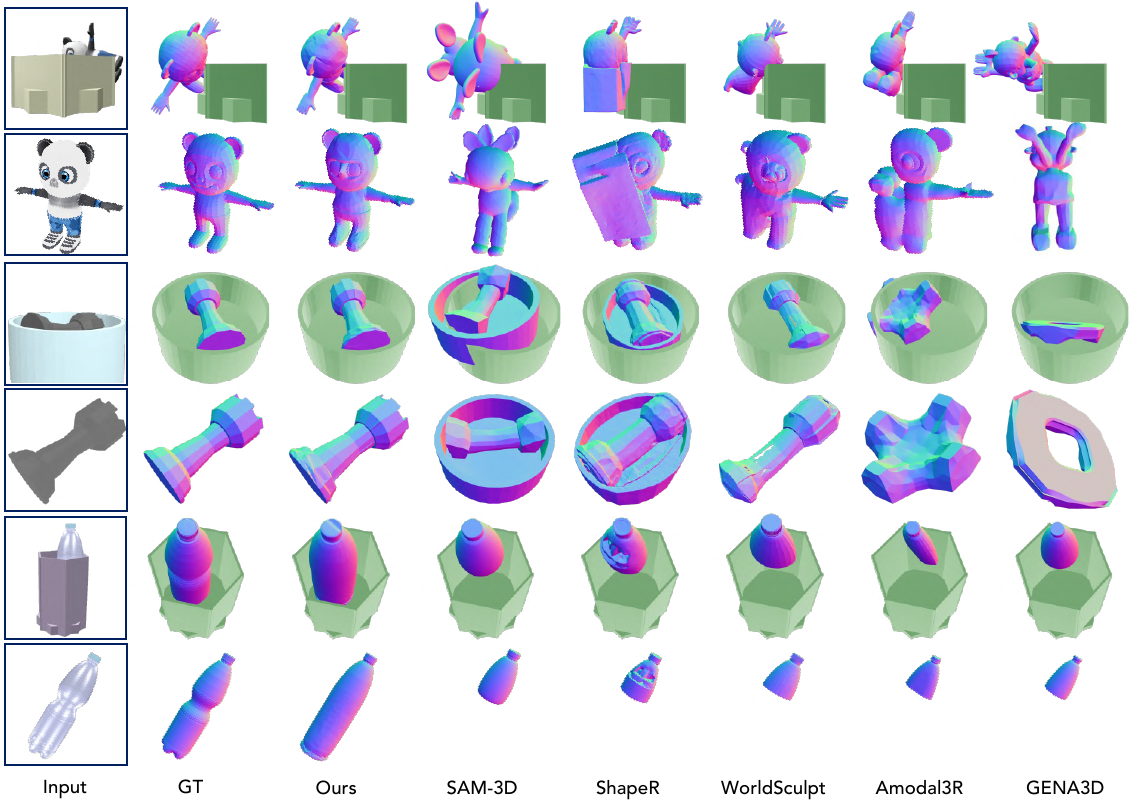}
    \caption{\textbf{Qualitative comparison.}
    % We compare our \ours{} with scene generation and amodal generation baselines. 
    % 첫 row에는 위에 input scene image, 그리고 아래에는 target의 비교를 위해 complete image를 넣었다. 각 scene마다 첫 row는 scene space에서의 shape with pose이고, 두 번째 row는 각 Object를 같은 뷰에서 본거다. \ours{} achieves superior generation quality in both shape and pose and simultaneously recover plausible interactions with neighbors, even under occlusion.
    Input scene images (top) and complete target images (bottom) are shown for reference. For each scene, the top row shows reconstructed shapes and poses in scene space, and the bottom row shows individual objects from a common viewpoint.
    }
    \label{fig:main-qualitative-comparison}
\end{figure*}

\begin{figure*}[!t]
    \centering
    \includegraphics[width=\linewidth]{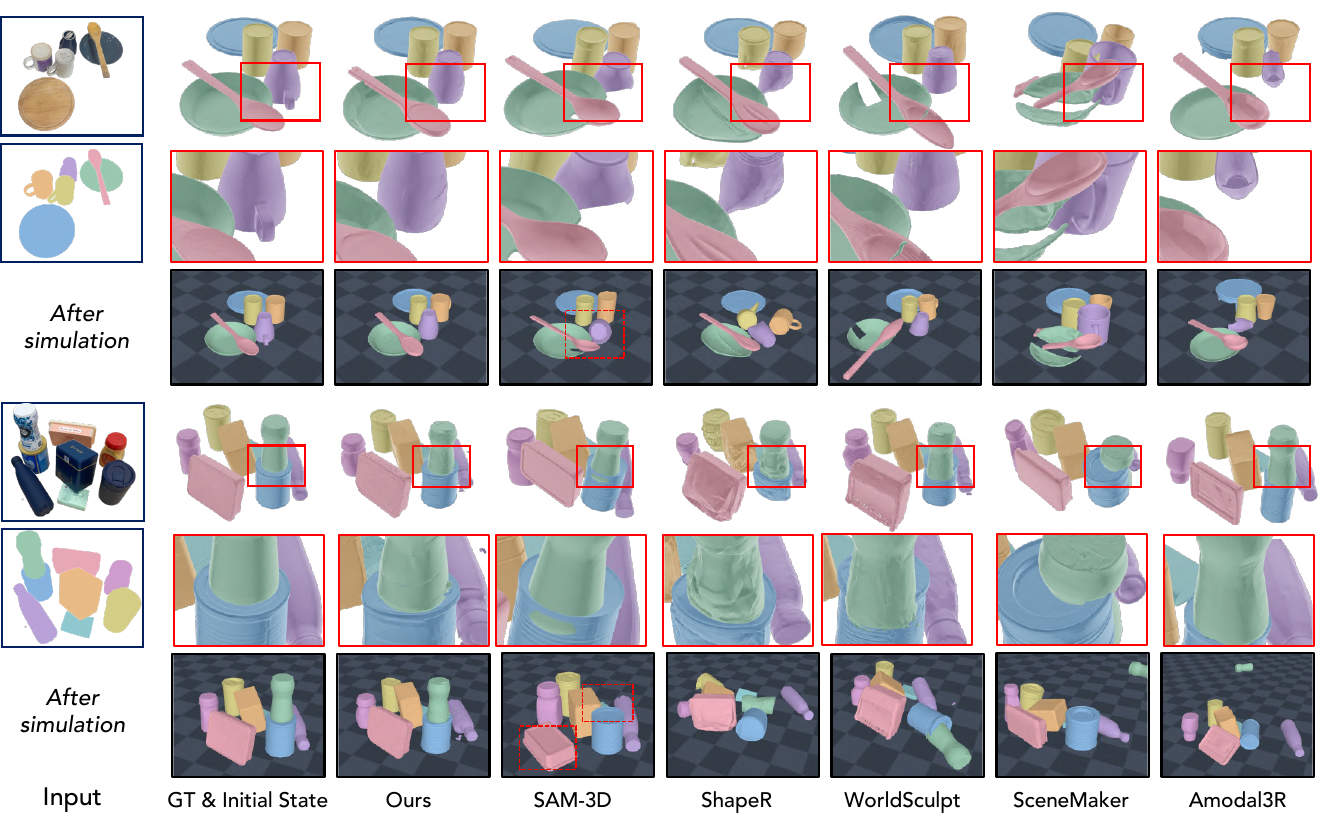}
    \caption{\textbf{Qualitative comparison.}
    The first column shows the scene image (top) and per-object masks (bottom), with colors matching the corresponding objects in the generated scenes. For each scene, we show the reconstruction and its final state after physics simulation~\citep{todorov2012mujoco}.
    }
    \label{fig:main-qualitative-comparison-scene}
\end{figure*}

\subsection{Comparisons}

\paragraph{Quantitative Comparison.}
As shown in Tab.~\ref{tab:toys4k_main_comparison}, \ours{} outperforms both scene generation baselines and amodal generation methods combined with pose estimation across reconstruction quality, generative quality, and physical stability. These results indicate that \ours{} recovers plausible shapes and well-aligned poses under occlusion, by leveraging surrounding spatial context to produce physically consistent scenes. Moreover, as shown in Tab.~\ref{tab:scene_comparison}, \ours{} maintains superior performance on real-world scene benchmarks with complex interactions and contacts, demonstrating its generalization beyond synthetic scenes. Across these comparisons, \ours{} achieves substantial improvements in physical stability including PD and \(D_{\mathrm{mean}}\), indicating
the effectiveness of conditioning object generation on neighboring geometry and physical relationships.

\paragraph{Qualitative Comparison.}
We show qualitative comparisons on Toys4K in Fig.~\ref{fig:main-qualitative-comparison} and on MessyKitchens~\citep{ansari2026messykitchens} in Fig.~\ref{fig:main-qualitative-comparison-scene}. As shown in Fig.~\ref{fig:main-qualitative-comparison}, \ours{} reconstructs plausible object shapes and poses even under occlusion, while better preserving the physical interactions with surrounding primitive objects without penetration. Fig.~\ref{fig:main-qualitative-comparison-scene} further shows that our method generalizes well to real scene images, producing more plausible reconstructions than the baselines and remaining stable after physics simulation. Note that Fig.~\ref{fig:main-qualitative-comparison-scene} results are obtained by autoregressive generation using estimated depth and VLM-based physical relation reasoning.

\subsection{Ablation Study}
We analyze the effects of the interaction conditioning signal and the \emph{visible-to-invisible (V2I) attention} described in Sec.~\ref{Method:v2iv-attn}. Further training details for the ablation study are provided in Appendix~\ref{Appendix:implementation-detail}.

\begin{figure*}[!t]
    \centering

    % Figure 1
    \begin{minipage}[t]{0.52\textwidth}
        \centering

        \includegraphics[width=\linewidth]{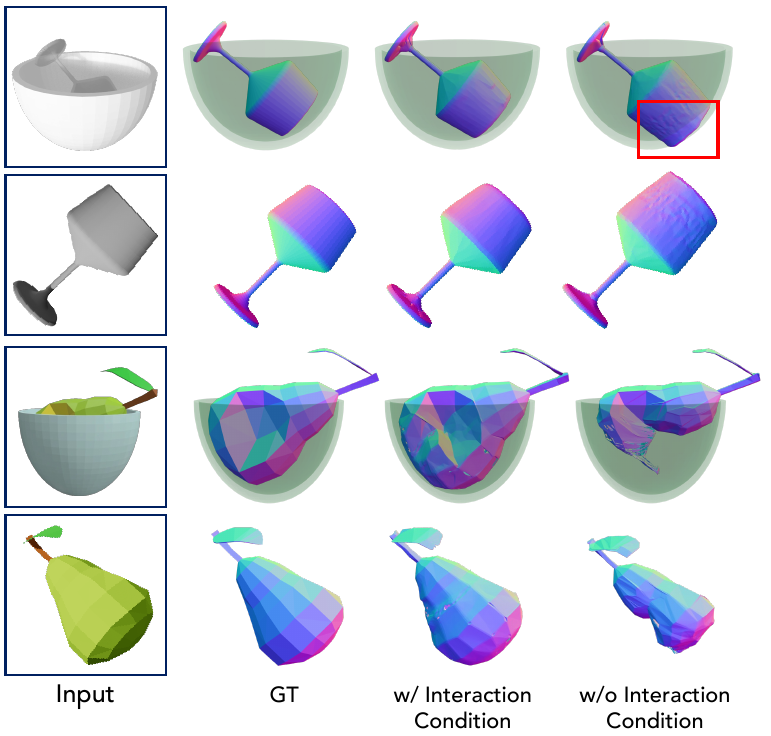}

        \captionof{figure}{\textbf{Qualitative comparison on interaction conditioning.} 
        % Interaction conditioning not only reduces interpenetration (top row) but also help complete the geometry of occluded region (bottom row).
        }
        \label{fig:ablation-qual-interaction}
    \end{minipage}
    \hfill
    % Figure 2
    \begin{minipage}[t]{0.47\textwidth}
        \centering

        \includegraphics[width=\linewidth]{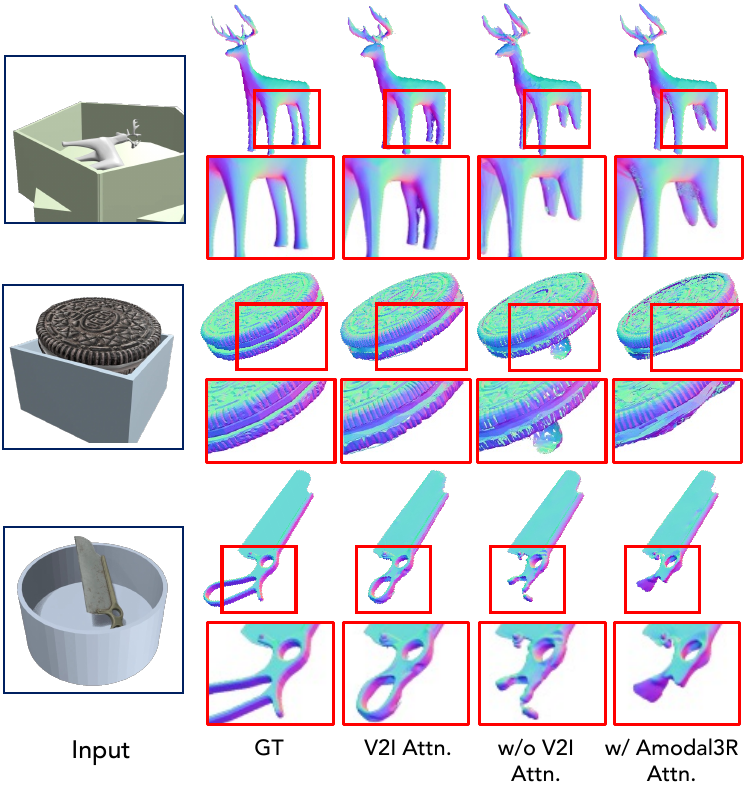}

        \captionof{figure}{\textbf{Qualitative comparison of amodal generation.}}
        \label{fig:ablation-qual-amodal}
    \end{minipage}

\end{figure*}

% For the ablation study, each variant is trained for fewer iterations than the full model. Further training details are provided in Appendix~\ref{Appendix:implementation-detail}.
\paragraph{Analysis on Interaction Conditioning.}
To assess the effect of interaction conditioning, we compare against a variant of our model without the interaction condition. As shown in Tab.~\ref{tab:abl_interaction} and Fig.~\ref{fig:ablation-qual-interaction}, this variant exhibits lower shape generation quality and reduced physical stability. In particular, penetration depth (PD) increases substantially, indicating more severe interpenetration between objects. These results suggest that interaction conditioning provides important cues not only for accurate reconstruction but also for generating physically plausible object configurations.

% ── Ablation: Interaction and amodal conditioning ─────────────
\begin{table*}[t]
    \centering
    \caption{
        \textbf{Ablation study on interaction conditioning and amodal generation.}
        We evaluate the effect of interaction conditions and architectural designs
        for amodal generation on Toys4K~\citep{stojanov2021using}.
        For \textit{w/ Amodal3R Attn}, we replace our \emph{V2I attention}
        with the occlusion-aware attention module of
        Amodal3R~\citep{wu2025amodal3r}.
    }
    \label{tab:abl_interaction}

    \small
    \setlength{\tabcolsep}{3pt}
    \renewcommand{\arraystretch}{1.15}

    \resizebox{\textwidth}{!}{%
    % Configuration (1) + Reconstruction (6) + Generation (5)
    % + Physical Stability (3) = 15 columns
    \begin{tabular}{@{}l*{14}{c}@{}}
        \toprule

        % Header 1: Metric groups
        \multirow{2}{*}{\textbf{Configuration}}
        & \multicolumn{6}{c}{\textbf{Reconstruction Quality}}
        & \multicolumn{5}{c}{\textbf{Generation Quality}}
        & \multicolumn{3}{c}{\textbf{Physical Stability}} \\

        \cmidrule(lr){2-7}
        \cmidrule(lr){8-12}
        \cmidrule(lr){13-15}

        % Header 2: Individual metrics
        % Reconstruction
        & CD-S~\(\downarrow\)
        & CD-O~\(\downarrow\)
        & F1-S~\(\uparrow\)
        & F1-O~\(\uparrow\)
        & IoU-B~\(\uparrow\)
        & ICP-Rot~\(\downarrow\)

        % Generation
        & MMD~\(\downarrow\)
        & COV~\(\uparrow\)
        & P-FID~\(\downarrow\)
        & Uni3D~\(\uparrow\)
        & ULIP~\(\uparrow\)

        % Physical stability
        & PD~\(\downarrow\)
        & \(D_{\mathrm{mean}}\)~\(\downarrow\)
        & \(E_{\mathrm{peak}}\)~\(\downarrow\) \\

        \midrule

        \textbf{Full model (Ours)}
        & \textbf{3.90} & \textbf{1.28} & \textbf{0.7963} & \textbf{0.7691} & \textbf{0.7603} & \textbf{5.50} & \textbf{1.563} & \textbf{84.3} & \textbf{1.875} & \textbf{0.6704} & \textbf{0.7069} & \textbf{0.0192} & \textbf{39.22} & \textbf{0.1193}  \\

        \midrule

        \quad w/o Interaction Cond.
        & 8.56 & 1.37 & 0.7599 & 0.7134 & 0.6918 & 8.66 & 1.841 & 80.66 & 1.938 & 0.6385 & 0.6766 & 0.4980 & 69.24 & 0.3240 \\

        \quad w/o V2I Attn.
        & 4.46 & 1.37 & 0.7905 & 0.7591 & 0.7506 & 6.00 & 1.629 & 84.01 & 1.949 & 0.6474 & 0.6897 & 0.0270 & 42.04 & 0.1396 \\

        \quad w/ Amodal3R Attn.
        & 4.29 & 1.34 & 0.7909 & 0.7659 & 0.7496 & 5.82 & 1.573 & 84.25 & 1.951 & 0.6615 & 0.7018 & 0.0281 & 40.67 & 0.1322 \\

        \bottomrule
    \end{tabular}%
    }
\end{table*}

% \begin{figure*}[!t]
%     \centering
%     \fbox{%
%         \parbox[c][0.3\textheight][c]{%
%             \dimexpr\textwidth-2\fboxsep-2\fboxrule\relax
%         }{%
%             \centering
%             \Large Ablation Qual Comparison
%         }%
%     }
%     \caption{Ablation Qual Comparison.}
%     \label{Figure:ablation-qual}
% \end{figure*}

% \captionof{figure}{\textbf{Effect of interaction conditioning.} Interaction conditioning not only reduces interpenetration but also guides the completion of occluded geometry.}

% \captionof{figure}{\textbf{Qualitative comparison of amodal generation.} V2I attention improves the completion of occluded regions.}
\paragraph{Attention Design for Amodal Generation.}
% We compare different architectural designs for amodal generation to evaluate the effect of the \emph{visible-to-invisible (V2I) attention} introduced in Sec.~\ref{Method:v2iv-attn}. As shown in Tab.~\ref{tab:abl_interaction}, removing the \emph{V2I} attention leads to 더 낮은 생성 및 reconstruction 성능을. Replacing it with the occlusion-aware attention module proposed in Amodal3R~\citep{wu2025amodal3r} results in 약간 향상되었지만 좀 덜 좋은. These results suggest that explicitly aggregating information from visible tokens provides effective guidance for completing invisible regions under occlusion.
We compare different architectural designs for amodal generation to evaluate the effect of the \emph{visible-to-invisible (V2I) attention} introduced in Sec.~\ref{Method:v2iv-attn}.
As shown in Tab.~\ref{tab:abl_interaction}, removing \emph{V2I attention} reduces both generative and reconstruction quality.
Replacing it with the occlusion-aware attention module of Amodal3R~\citep{wu2025amodal3r} yields a modest improvement over the variant without V2I attention, but performance remains below that of our design.
These results suggest that explicitly aggregating information from visible tokens provides effective guidance for completing invisible regions under occlusion.
\section{Conclusion}
% We propose \ours{}, a generative framework for single-image 3D scene reconstruction that recovers objects which are physically and geometrically coherent as a scene. The key component is an object-interaction conditioning scheme that explicitly incorporates neighboring geometry and physical relations into generation. We further introduced \textit{ComOb}, a large-scale simulation-based dataset containing diverse objects in physically stable configurations with explicit interaction annotations. Across synthetic and real-world benchmarks, our framework produces more accurate and physically consistent scene reconstructions, particularly in challenging regions occluded by interacting objects. We believe this approach provides a practical step toward versatile compositional 3D scene generation.
We propose \ours{}, a generative framework for single-image 3D scene reconstruction that explicitly conditions object generation on neighboring geometry and physical relations. We also introduce \textit{ComOb}, a large-scale simulation-based dataset of diverse objects in physically stable configurations with interaction annotations. Experiments on synthetic and real-world benchmarks demonstrate improved reconstruction quality and physical consistency, even when interacting regions are occluded. We believe \ours{} provides a practical step toward versatile compositional 3D scene generation.

\subsection*{AI use statement}

% (This section is \textbf{required} and does not count toward the page limit.)
% In this work, we used generative AI tools for assisting with code implementation, manuscript writing and the preparation of the result visualizations. We did not use generative AI tools for research ideation, model and experimental design, or analysis. These tasks were performed by the authors. We reviewed all AI-assisted work. Specifically, we checked that the code functioned as intended, reviewed the text for technical accuracy and consistency with our intended meaning, and verified that the visualizations accurately represented our methods and experimental results. We take responsibility for the final content of this work, including text, claims, and artifacts produced with the aid of generative AI.

In this work, we used generative AI tools for polishing the writing of the manuscript and the preparation of the result visualizations. We did not use generative AI tools for research ideation, model and experimental design, or analysis. These tasks were performed by the authors. We reviewed all AI-assisted work. Specifically, we reviewed the text for technical accuracy and consistency with our intended meaning, and verified that the visualizations accurately represented our methods and experimental results. We take responsibility for the final content of this work, including text, claims, and artifacts produced with the aid of generative AI.

\bibliography{iclr2027_conference}
\bibliographystyle{iclr2027_conference}

\clearpage
\appendix
\startcontents[appendix]

{\LARGE\bfseries Appendix\par}
\vspace{1.5em}

\appendixtoc
\section{Related Works}
\paragraph{3D generative models.}
With large-scale 3D datasets~\citep{deitke2023objaverse,deitke2023objaversexl}, 3D generation has shifted toward 3D-native models over compact latents: one line encodes a shape as an unordered set of latent vectors~\citep{zhang20233dshape2vecset,zhang2024clay,li2025triposg,zhao2025hunyuan3d}, while another line attaches latents to a sparse voxel grid and decodes them into 3D assets~\citep{xiang2025structured, wu2026direct3d, xiang2026native}. Across both lines, the input image feature is typically injected via cross-attention into a canonical-pose generator~\citep{cheng2023sdfusion,hong2024lrm,li2025triposg,chen2026sam}. A complementary line instead grounds the condition in the generation space itself: Pixal3D~\citep{li2026pixal3d} back-projects pixel features along camera rays and generates in the input camera frame, while 3D priors such as point cloud~\citep{xia2026points, siddiqui2026shaper} are also embedded into the latent rather than attended to. These models, however, assume a clean image of a single object, whereas objects in real scenes occlude one another. Amodal completion addresses this in image space~\citep{ozguroglu2024pix2gestalt,ao2025open,qu2026deocc}, and recent works complete occluded objects in the 3D latent space, conditioning attention on visibility and occlusion cues~\citep{wu2025amodal3r,hu2026axolotl3d,zhou2025gena3d}. All of these, however, complete each object from its own visible region alone, without leveraging cues from the surrounding scene. Motivated by this, we condition generation on the spatial context and physical relations of neighboring objects, and complete occluded regions by attending from hidden to visible tokens.

\paragraph{3D scene generation.}
Scenes have been built either by retrieving assets from offline libraries~\citep{dai2024automated}, which limits open-set diversity, or by learning scene-native generative models~\citep{liu2022towards,dahnert2024coherent,tang2024diffuscene} from scene datasets~\citep{fu20213d,dai2017scannet,azinovic2022neural}, which confines them to specific domains such as indoor rooms. Object-native pipelines~\citep{ardelean2025gen3dsr,zhao2025depr,chen2026sam} lift this restriction with generators trained on open-set data, generating each object before registering it into the scene. Concurrently, WorldSculpt~\citep{niu2026worldsculpt} extends Pixal3D~\citep{li2026pixal3d} to multi-view setting for scene-level generation. Since objects are generated independently, the composed scenes often exhibit physically implausible artifacts such as penetration or floating. To enforce physical plausibility, one line resolves these violations at test time, using VLM relation graphs to refine poses~\citep{yao2025cast,li2026phi}, or running a simulation in the loop~\citep{lin2026pat3d,lee2026simuscene}, while another builds physics into the generation process~\citep{li2026spread, chen2026physcene3d}. More recently, several methods generate multiple instances in a single pass, coupling shape and pose through attention~\citep{huang2025midi,meng2026scenegen,lin2026partcrafter,zhang2026pixarmesh} or autoregressive co-generation~\citep{tang2026co}, while another line~\citep{shi2026scenemaker, ansari2026messykitchens} predicts object poses jointly, letting attention among them settle into a layout. In contrast to this family, where object interactions remain implicit in attention or confined to pose, we explicitly condition generation on the spatial context and physical relations between objects.

% More recently, a feed-forward family generates multiple instances in a single pass, coupling shape and pose through attention~\citep{huang2025midi,meng2026scenegen,lin2026partcrafter,zhang2026pixarmesh} or autoregressive co-generation~\citep{tang2026co}, while another line~\cite{shi2026scenemaker, ansari2026messykitchens} predicts the poses of all objects jointly, letting attention among them settle into a coherent layout. In contrast to this family, where object interaction remain implicit in attention or confined to pose, we explicitly condition the generation on the spatial context and physical relations between objects.

\section{Implementation Details} \label{Appendix:implementation-detail}

\subsection{Model Details}

\paragraph{Overall Model Architecture.}
We build \ours{} on TRELLIS.2~\citep{xiang2026native}, which consists of a 30-block DiT backbone for each generation stage. With the added visible-to-invisible attention layers, the sparse structure and structured latent generation models contain approximately 1.83B and 1.68B parameters, respectively. Each block has a lightweight encoder for the per-token nearest-surface condition (Sec.~\ref{Method:per-token-injection}). 
% The conditoin embeddings are masked, concatenated, and fused by a two-layer MLP mapping \(\mathbb{R}^{256}\) to \(\mathbb{R}^{1536}\).

\paragraph{Image Condition Masking.}
Following prior works~\citep{wu2025amodal3r, zhou2025gena3d, hu2026axolotl3d}, we blend each DINOv3~\citep{simeoni2025dinov3} patch feature \(\mathbf{f}\) with a learnable null embedding \(\mathbf{e}_{\mathrm{empty}}\) according to its occlusion ratio \(\rho\in[0,1]\):
\begin{equation}
\tilde{\mathbf{f}} = (1-\rho)\mathbf{f} + \rho\,\mathbf{e}_{\mathrm{empty}}.
\end{equation}
The same masking is applied to the projected image condition \(c_{\mathrm{img}}\) defined in Eqn.~\ref{equ:pose-alignment}.

\paragraph{Visible-to-Invisible Attention.}
Within each sample, tokens with fully observed projected features are classified as visible, while those receiving any null contribution are classified as invisible. Tokens projecting outside the image crop are excluded from this attention branch. In each DiT block, we initialize the weights of the branch's normalization and query/output projections from the pretrained cross-attention, and its key/value projections from self-attention. The residual output is applied only to invisible tokens through a zero-initialized gate, so the added branch initially leaves the original block update unchanged. When no invisible tokens are present, this layer is skipped.

\subsection{Training Details}
We initialize the weights of TRELLIS.2 backbone and the back-projected image-feature conditioning components of \ours{} from Pixal3D~\citep{li2026pixal3d}, and train the model on our ComOb dataset described in Sec.~\ref{Method:comob}. We detail the dataset construction in Sec.\ref{Appendix:dataset-details}. We train the sparse structure generation model for 200K iterations and the structured latent generation model for 100K iterations. For both training, we use the AdamW optimizer~\citep{loshchilov2017decoupled} with a learning rate of \(1\times10^{-4}\) and a weight decay of \(0.01\). All models are trained on two NVIDIA H200 GPUs with a batch size of 16. Training takes approximately 7 days for the sparse structure DiT and 9 days for the structured latent DiT. Note that, for the ablation study, we train the sparse structure generation model for 100K iterations, while keeping the structured latent generation model unchanged. We also apply condition dropout for classifier-free guidance~\citep{ho2022classifier}. All conditions are jointly dropped with probability \(0.1\), while individual condition groups are independently dropped with probability \(0.05\). For sparse structure generation, these groups consist of the image features, depth condition, neighboring-surface geometry, and relation condition; for structured latent generation, they consist of the image, geometry, and relation conditions. 

\paragraph{Curriculum Training with Occlusion.} To improve robustness to diverse forms of occlusion while preserving the prior of the pretrained shape generator, we use three types of training images: complete images without occlusion, images occluded by the primitives, and images with synthetic occlusion. For synthetic occlusion, we follow the protocol of Axolotl3D~\citep{hu2026axolotl3d}, compositing synthetic occluders onto complete images. Following~\citep{siddiqui2026shaper}, we adopt a two-stage curriculum training strategy. In Stage~1, the model is trained only on complete images with all conditions provided. In Stage~2, we additionally include primitive-occluded and synthetically occluded images, sampling the three image types at a ratio of \(1{:}2{:}1\), respectively. Specifically, we train the sparse structure generation model for 50K iterations in Stage~1 and 150K iterations in Stage~2. For structured latent generation, we train for 30K and 70K iterations in Stage~1 and Stage~2, respectively. For the ablation study, we train 50K iterations in Stage~2 for the sparse structure generation model while keeping Stage~1 training iterations unchanged. The ablation study is conducted with models trained for the same number of iterations to exclude the effect of training configuration.

\paragraph{Depth Augmentation.} To improve robustness to noisy depth estimates at inference time, we augment the depth condition during training. For sparse structure generation, with probability \(0.3\), we perturb observed depth points along their camera rays with Gaussian noise and re-voxelize them within a local neighborhood. We further randomly drop \(10\%\) of the observed depth voxels. These perturbations are applied only to the depth condition, leaving the image features unchanged.

\begin{figure}[t]
    \centering
    \includegraphics[width=\linewidth]{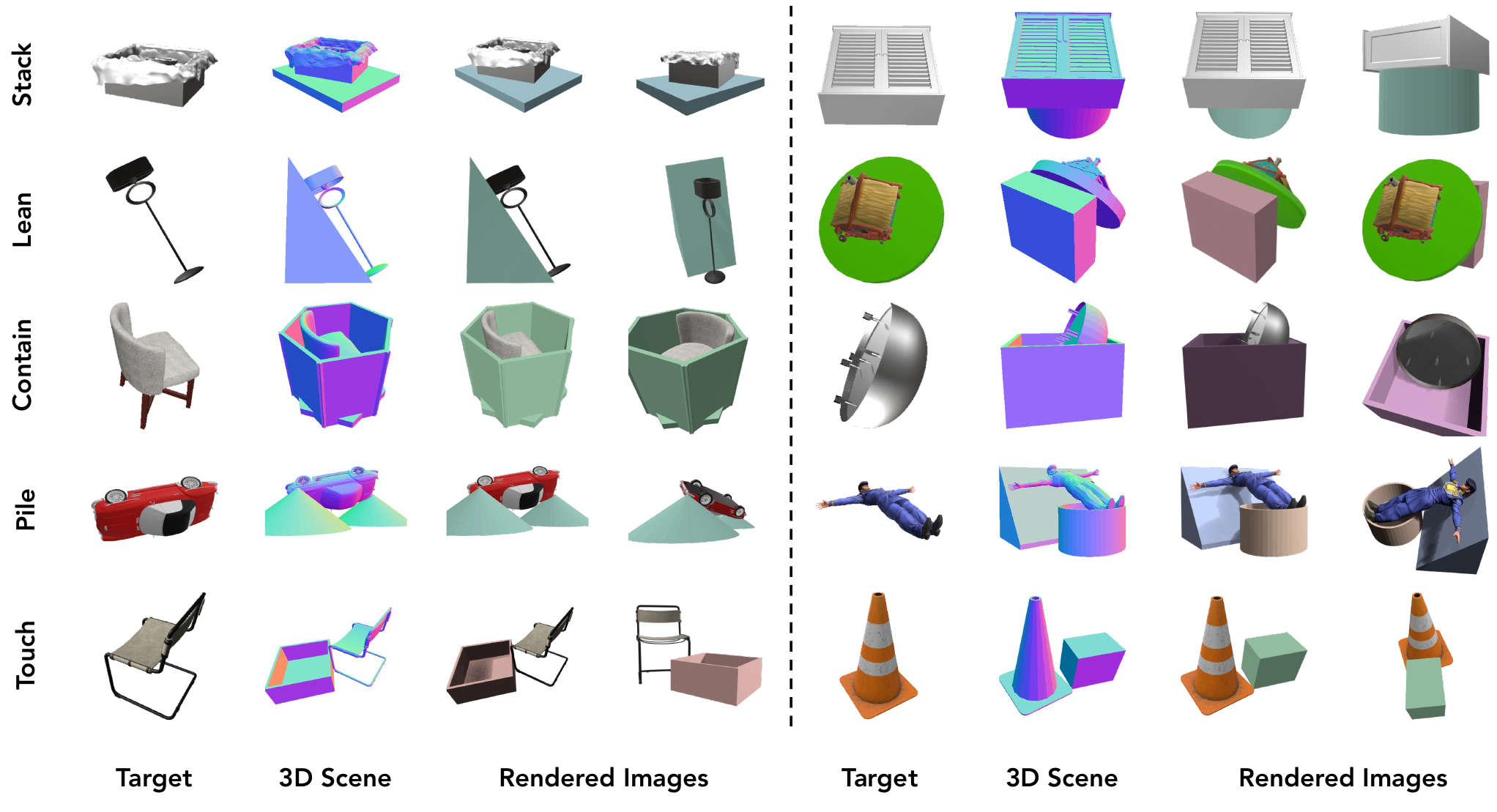}
    \caption{\textbf{Examples of \textit{ComOb}.}
    We showcase scenes from the \textit{ComOb} dataset. Each row corresponds to a different interaction type. From left to right, we show the complete target-object image, the normal map of the 3D scene, and images rendered from viewpoints used during training.
    }
    \label{figure-comob-example}
\end{figure}

\subsection{Dataset Generator Details} \label{Appendix:dataset-details}
We construct our dataset, \textbf{ComOb}, to provide physically consistent scenes with explicit object-interaction annotations. Each scene contains a single target object and one or more neighboring primitives. We choose and size the primitives according to the target geometry and intended interaction. We then run the simulation engine with the target while keeping the primitives fixed, and retain only scenes that pass the relation and geometry checks described below. Using approximately 400K unique target objects from TRELLIS-500K~\citep{xiang2025structured}, we generate 1.2M scenes. Examples of the generated scenes are shown in Fig.~\ref{figure-comob-example}.

\paragraph{Primitive Selection.}
We use twelve parametric primitives: \textit{boxes}, \textit{planks}, \textit{cylinders}, \textit{cones}, \textit{spheres}, \textit{wedges}, \textit{tubes}, \textit{tori}, \textit{cups}, \textit{open bins}, \textit{polygonal bins}, and \textit{bowls}. The available primitive types depend on the intended interaction:
\begin{itemize}
    \item \textbf{Stacking:} boxes, planks, and cylinders.
    \item \textbf{Containment:} cups, open bins, polygonal bins, tubes, and bowls.
    \item \textbf{Leaning and touching:} all twelve primitive types.
    \item \textbf{Piling:} cylinders, tubes, cones, wedges, boxes, and spheres.
\end{itemize}

\paragraph{Primitive Sizing.}
% We size each primitive relative to the target geometry. For interactions other than containment, we sample the primitive's longest dimension within a factor of two of the target's, allowing the primitive to be either larger or smaller. For containment, the primitive size is instead determined by the target diameter relative to the container opening. Across all interaction types, we constrain the primitive-to-target size ratio to remain between \(1/2\) and \(2\).
We sample the primitive's longest dimension between half and twice that of the target object. For containment, we instead determine the primitive size based on the target diameter relative to the container opening, while enforcing the same bounds on its longest dimension.

\paragraph{Interaction-specific Placement.}
We generate scenes with 5 predefined interactions: \{\textit{stack}, \textit{lean}, \textit{contain}, \textit{pile}, \textit{touch}\}. For each interaction type, we initialize each scene according to the intended interaction and check the resulting configuration after simulation:
\begin{itemize}
    \item \textbf{Stack.}
    We release the target slightly above the supporting surface with a small horizontal perturbation. After settling, at least half of its footprint must overlap the support.
    \item \textbf{Lean.}
    We place a primitive on the side toward which the target is expected to topple. This direction is estimated from the projected center of mass relative to the center of its ground-support region. After settling, we remove the primitive in a separate simulation. The scene is accepted as leaning only if the target falls without this support.

    \item \textbf{Contain.}
    We vary the target diameter relative to the container opening to obtain roomy, near-fit, and oversize configurations. Roomy targets are released near the cavity floor; the others are dropped just above the rim. We retain stable configurations in which the target is fully contained, partially inserted, or supported by the rim.
    \item \textbf{Pile.}
    We arrange two to four primitives, potentially of different types, in an inward-facing ring that forms a concave pocket. We drop the target into its center and discard configurations in which it falls out.
    \item \textbf{Touch.}
    We place the target beside a primitive with a small gap. Before the final stability check, we reduce this gap to establish contact. The resulting configuration must satisfy the penetration tolerance.
\end{itemize}

\paragraph{Physical Settling.}
We simulate each candidate scene in MuJoCo~\citep{todorov2012mujoco}, gradually reducing damping as the target settles. Containment and piling receive longer settling periods than the other interactions. We retain a candidate only if its linear and angular speeds remain below \(0.025\,\mathrm{m/s}\) and \(0.12\,\mathrm{rad/s}\), respectively, throughout the final observation window.

\paragraph{Relation Annotations.}
We assign each accepted target--primitive pair a relation \(r\in\mathcal{R}\), following Sec.~\ref{Method:object-interaction}. The scene construction type does not necessarily become the pairwise relation label. In particular, \textit{pile} describes how a scene is constructed, not an additional relation. Within a piling scene, primitives that contact the target are labeled \textit{stack}, while the remaining primitives are labeled \textit{none}.

\paragraph{Rendering Details.}
We render each settled scene in Blender using the EEVEE rasterizer, with the field of view randomly sampled from \([35^\circ,70^\circ]\). To obtain a suitable level of occlusion, we generate candidate viewpoints over a predefined set of azimuth and elevation angles and retain those whose target-object visible ratio lies within \([0.3,0.9]\). Other rendering configurations, including lighting, follow TRELLIS~\citep{xiang2025structured}.
% after the objects are placed according to the specified interaction, the scene is settled through MuJoCo simulation~\citep{todorov2012mujoco} to obtain a physically consistent configuration. Scenes that no longer satisfy the specified interaction after settling are filtered out. 

\subsection{Inference Pipeline Details} \label{Appendix:inference-details}

\paragraph{Data Preprocessing.}
At inference, our framework takes a monocular RGB image together with a depth map, camera parameters, and per-object segmentation masks. We obtain the depth map and camera parameters using the off-the-shelf depth estimation model MoGe3~\citep{kong2026moge}, and extract object masks with SAM3~\citep{carion2026sam}. To automatically identify the objects to segment, we use Qwen3-VL-30B~\citep{bai2025qwen3} to extract the distinct object instances present in the input image. For each target object \(o_i\) with segmentation mask \(M_i\), we define its occlusion mask as the union of the masks of all other objects:
\begin{equation}
M_i^{\mathrm{occ}} =  \bigcup_{j\neq i} M_j.
\end{equation}
By defining occlusion in this way, we do not assume or estimate the amodal extent of the target object in the image; instead, all regions occupied by other objects are treated as potential occlusions of the target.

% \input{figures/figure-inference-pipeline}
% \input{tables/pseudocode_appendix}
% \paragraph{Physical Dependency Graph Construction.}
% Following prior works~\citep{yao2025cast, li2026phi, chen2026physcene3d}, we employ a VLM to construct a directed graph based on physical dependency relations. Our procedure consists of a two-stage prompting pipeline designed to improve spatial grounding and reduce ambiguity in physical reasoning. In the first stage, we provide Qwen3-VL-30B~\citep{bai2025qwen3} with the input image and a colored segmentation map to extract object metadata that associates each object with its visual attributes and unique RGB color in the segmentation map. This establishes a consistent correspondence between textual object labels and their image regions. In the second stage, the resulting metadata, together with the original image and segmentation map, is provided to the VLM to infer directed support relations between objects. We provide the prompt examples for each stage in Prompt ~\ref{Appendix_prompt_A} and ~\ref{Appendix_prompt_B}.

\paragraph{Physical Dependency Graph Construction.}

\begin{wrapfigure}{r}{0.54\textwidth}
    \vspace{-0.5em}

    % =========================================================
    % Top rule
    % =========================================================
    \hrule height 0.8pt
    \vspace{3pt}

    % =========================================================
    % Algorithm title
    % =========================================================
    \refstepcounter{algorithm}
    \noindent
    {\small
    \textbf{Algorithm~\thealgorithm: Confidence-Weighted Physical Ordering}}
    \label{alg:optimal-topological-sort}

    \vspace{3pt}

    % Rule below title
    \hrule height 0.4pt
    \vspace{3pt}

    % =========================================================
    % Pseudocode
    % =========================================================
    \footnotesize
    \renewcommand{\algorithmicrequire}{\textbf{Input:}}
    \renewcommand{\algorithmicensure}{\textbf{Output:}}
    \begin{algorithmic}[1]

    \Require Objects \(\mathcal{V}\), ground root \(r\),
    $\mathcal{E}_d\!=\!\{(v_j\!\to\! v_i,w)\}$ with confidence $w$,
    %directed relations \(\mathcal{E}_d\),    
    undirected relations \(\mathcal{E}_u\)

    \Ensure Generation order \(\pi\)

    % \Statex \hspace{-1em}$\mathrm{parent}[S]$: the last object added on the best path to S.

    \State \(\mathcal{N}(v) \gets\) vertices directly reachable from \(v\) by
    \(\mathcal{E}_d\) or \(\mathcal{E}_u\)

\State $\mathscr{S}\gets
\{S\subseteq\mathcal{V}\mid r\in S\}$

    % \State \(D[\{r\}] \gets 0\), and \(D[S] \gets \infty\) for all other \(S\)
    \State $D[S]\gets\infty\ \ \forall S\in\mathscr{S}$; \ \ $D[\{r\}]\gets 0$

    \ForAll{\(S\in\mathscr{S}\) in increasing order of \(|S|\)}

        \ForAll{\(v \in \mathcal{V}\setminus S\) such that
        \(\mathcal{N}(v)\cap S\neq\varnothing\)}

            % \State \(c \gets 0\)

            \State
            \(c(v, S) \gets 
            \displaystyle
            \sum_{\substack{
                (v\!\to\! u,w)\in\mathcal{E}_d\\
                u\in S
            }} w\)

            \If{\(D[S] + c (v,S)< D[S\cup\{v\}]\)}

                \State
                \(D[S\cup\{v\}]
                \gets D[S] + c(v,S)\)

                \State
                \(\mathrm{parent}[S\cup\{v\}]
                \gets (S,v)\)

            \EndIf

        \EndFor

    \EndFor

    \State
    \(\pi \gets\) backtrack from
    \(\mathcal{V}\) to \(\{r\}\)
    using \(\mathrm{parent}\)

    \State \Return \(\pi\)

    \end{algorithmic}

    % =========================================================
    % Bottom rule
    % =========================================================
    \vspace{3pt}
    \hrule height 0.8pt

    \vspace{-0.5em}
\end{wrapfigure}
Following prior works~\citep{yao2025cast, li2026phi, chen2026physcene3d}, we employ a VLM to construct a graph based on physical relations among objects. Our procedure consists of a two-stage prompting pipeline designed to improve spatial grounding and reduce ambiguity in physical reasoning. In the first stage, we provide Qwen3-VL-30B~\citep{bai2025qwen3} with the input image and a colored segmentation map to extract object metadata that associates each object with its visual attributes and unique RGB color in the segmentation map. This establishes a consistent correspondence between each object and its image region. In the second stage, the resulting metadata, together with the original image and segmentation map, is provided to the VLM to infer physical relations with confidence between objects. We provide the prompt examples for each stage in Prompt~\ref{Appendix_prompt_A} and~\ref{Appendix_prompt_B}.

%The floor is treated as an independent root node, yielding a bottom-up generation process grounded in the scene support structure.
Specifically, we distinguish directed physical dependencies from undirected contact relations. A directed edge \(o_j \rightarrow o_i\) indicates that \(o_i\) depends on \(o_j\) for its physical configuration, and thus \(o_j\) should be generated before \(o_i\). In contrast, a \textit{touch} relation is treated as undirected. It does not prescribe an ordering, but requires at least one interacting neighbor to be available when the object is generated. Since VLM predictions may contain conflicting or cyclic dependencies, directly applying a topological sort does not always yield a valid generation order. We therefore employ a confidence-weighted ordering procedure, described in Alg.~\ref{alg:optimal-topological-sort}. The algorithm incrementally adds objects that have at least one physically related object already generated. Among all feasible orders, it minimizes the total confidence of directed edges that conflict with the resulting order, thereby preferentially preserving high-confidence physical dependencies while resolving cyclic predictions. Directed edges inconsistent with the selected order are discarded, resulting in an acyclic dependency graph. The final autoregressive generation follows this optimized ordering. Consequently, each object is generated only after relevant neighboring geometry has become available for interaction conditioning, while undirected contacts can still constrain which objects are eligible to be generated next.

% Specifically, we distinguish directed physical dependencies from undirected contact relations. A directed edge \(o_j \rightarrow o_i\) indicates that \(o_i\) depends on \(o_j\) for its physical configuration, and thus \(o_j\) should be generated before \(o_i\). In contrast, a \textit{touch} relation is treated as undirected. It does not prescribe an ordering, but requires at least one interacting neighbor to be available when the object is generated. Since VLM predictions may contain conflicting or cyclic dependencies, directly applying a topological sort does not always yield a valid generation order. We therefore employ a confidence-weighted optimal topological sorting procedure, described in Alg.~\ref{alg:optimal-topological-sort}. Starting from the root node, the algorithm incrementally adds objects that have at least one physically related object already generated. Among all feasible orders, it minimizes the total confidence of directed edges that conflict with the resulting order, thereby preferentially preserving high-confidence physical dependencies while resolving cyclic predictions. Directed edges inconsistent with the selected order are discarded, resulting in an acyclic dependency graph. The final autoregressive generation order follows this optimized ordering. Consequently, each object is generated only after relevant neighboring geometry has become available for interaction conditioning, while undirected contacts can still constrain which objects are eligible to be generated next.

\section{Evaluation Details} \label{Appendix:evaluation-detail}
\subsection{Evaluation Protocol Details}

For evaluation, all methods receive the same scene image, per-object masks, and depth map. When estimated depth is used, we apply median-based alignment. We evaluate per-object meshes in a shared world coordinate frame, with object position, orientation, and scale preserved in the mesh vertices.
Since amodal generation baselines~\citep{wu2025amodal3r, zhou2025gena3d} generate objects in canonical object space, we register their outputs to the scene coordinate frame using an off-the-shelf pose estimator~\citep{wen2024foundationpose}. We evaluate reconstruction and generative quality using surface point clouds obtained through farthest point sampling (FPS).
For scene-level evaluation, including \(\text{CD-S}\), \(\text{F1-S}\), \(\text{IoU-B}\), and \(\text{ICP-Rot}\), we first align the predicted scene to the ground-truth scene using a single rigid ICP transformation estimated from the combined surface points of all objects. This alignment preserves the relative arrangement of objects within the scene. For object-level metrics, including \(\text{CD-O}\), \(\text{F1-O}\), and generative quality metrics, we normalize each object to a common range and perform per-object ICP alignment, so that evaluation focuses on individual shape rather than scene placement. For physical stability, following prior work~\citep{lee2026simuscene, li2026phi}, we simulate the reconstructed scenes in MuJoCo~\citep{todorov2012mujoco} and evaluate their stability during simulation. Penetration depth is measured in the initial configuration, before simulation.

\subsection{Evaluation Metric Details}
Throughout this section, let \(\hat{Q}\) and \(Q\) denote the predicted and ground-truth point sets, respectively, either for an individual object or for an entire scene obtained by concatenating the point sets of all objects. All point sets are sampled from the surface of the corresponding mesh.

\paragraph{Chamfer Distance (CD-S, CD-O).}
We measure the geometric discrepancy between \(\hat{Q}\) and \(Q\) using the symmetric mean of squared nearest-neighbor distances:
\begin{equation}
\mathrm{CD}(\hat{Q},Q)
=
\frac{1}{|\hat{Q}|}
\sum_{\hat{q}\in\hat{Q}}
\min_{q\in Q}
\|\hat{q}-q\|_2^2
+
\frac{1}{|Q|}
\sum_{q\in Q}
\min_{\hat{q}\in\hat{Q}}
\|q-\hat{q}\|_2^2.
\end{equation}
The first term measures the distance from the predicted geometry to the ground truth, while the second measures the distance from the ground truth to the prediction. We report CD-S at the scene level, where the point sets of all objects are concatenated, and CD-O at the individual-object level.

\paragraph{F-Score (F1-S, F1-O).}
Thresholding the same nearest-neighbour distances, this time Euclidean, at \(\tau\) gives a precision and a recall:
\begin{equation}
\mathrm{Prec}_{\tau} = \frac{1}{|P|}\sum_{p \in P}\mathbbm{1}\Big[\min_{q \in Q}\lVert p-q \rVert < \tau\Big],
\qquad
\mathrm{Rec}_{\tau} = \frac{1}{|Q|}\sum_{q \in Q}\mathbbm{1}\Big[\min_{p \in P}\lVert q-p \rVert < \tau\Big],
\end{equation}
and the F-Score is their harmonic mean. We use \(\tau = 0.005\), and report F1-S and F1-O at the scene and object level respectively.

\paragraph{Bounding-box IoU (IoU-B).}
We first align the predicted scene to the ground-truth scene using a single rigid ICP transformation. For each object, we then compute the volumetric intersection-over-union between the axis-aligned bounding boxes of the aligned predicted point set and the ground-truth point set.

\paragraph{Rotation Error (ICP-Rot).}
Let \(R \in \mathrm{SO}(3)\) denote the rotational component of the rigid transformation obtained by aligning a predicted object to its ground truth using ICP. We report the corresponding rotation angle,
\begin{equation}
\theta
=
\arccos
\left(
\frac{\operatorname{tr}(R)-1}{2}
\right),
\end{equation}
in degrees.

\paragraph{Penetration Depth (PD).}
We measure interpenetration before simulation as:
\begin{equation}
PD =
\frac{1}{N}
\sum_{i=1}^{N}
\sum_j
\max\left\{0,-\phi_j(\mathbf{x}_i)\right\},
\end{equation}
where \(\mathbf{x}_i\) is the world-space position of the \(i\)-th of \(N=20{,}480\) fixed, area-uniform target-surface samples, and \(\phi_j\) denotes the signed distance to neighboring object \(j\) at its initial pose, with negative values inside. Non-penetrating samples contribute zero.

% \paragraph{Penetration Depth.}
% To measure geometric interpenetration before and after simulation, we compute:
% \begin{equation}
% P_{\mathrm{mean}}(t)=\frac{1}{N} \sum_{i=1}^{N} \sum_j \max\left\{0, \phi_j\big(\mathbf{x}_i(t)\big)\right\},
% \qquad
% t\in\{0,T\},
% \end{equation}
% where \(\mathbf{x}_i(t)\) denotes the world-space position of the \(i\)-th of \(N=20{,}480\) fixed, area-uniform target-surface samples, and \(\phi_j\) is the signed distance function of the \(j\)-th watertight context primitive, with negative values indicating points inside the primitive. Non-penetrating samples therefore contribute zero.

\paragraph{Mean Displacement \(D_{\mathrm{mean}}\).}
To quantify the change in object pose after simulation, we measure the mean displacement of the same fixed target-surface samples:
\begin{equation}
D_{\mathrm{mean}} = \frac{1}{N} \sum_{i=1}^{N} \left\| \mathbf{x}_i(T)-\mathbf{x}_i(0) \right\|_2,
\end{equation}
where \(N=20{,}480\), \(\mathbf{x}_i(t)\) is the world-space position of the \(i\)-th surface sample, and \(T=300\,\mathrm{s}\) is the simulation duration.

\paragraph{Peak Kinetic Energy \(E_{\mathrm{peak}}\).}
To capture transient translational and rotational motion while reducing sensitivity to object mass, we report the peak kinetic energy per unit mass:
\begin{equation}
E_{\mathrm{peak}} = \max_{t\in\mathcal{T}} \frac{ m\|\mathbf{v}_{\mathrm{COM}}(t)\|_2^2 + \boldsymbol{\omega}(t)^\top \mathbf{I}_{\mathrm{COM}}(t) \boldsymbol{\omega}(t) }{ 2m },
\end{equation}
where \(m\) is the object mass, \(\mathbf{v}_{\mathrm{COM}}(t)\) is its center-of-mass velocity, \(\boldsymbol{\omega}(t)\) is its angular velocity, and \(\mathbf{I}_{\mathrm{COM}}(t)\) is the inertia tensor about the center of mass. The set \(\mathcal{T}\) contains simulation states sampled at \(1\,\mathrm{ms}\) intervals.

\paragraph{MMD and COV.}
We use Minimum Matching Distance (MMD) and Coverage (COV)~\citep{achlioptas2018learning} to compare the set of generated objects with the set of ground-truth objects, using Chamfer Distance as the underlying distance measure. Let \(\hat{\mathcal{Q}}=\{\hat{Q}\}\) and \(\mathcal{Q}=\{Q\}\) denote the sets of generated and ground-truth objects, respectively. MMD averages, over each ground-truth object, the distance to its nearest generated object:
\begin{equation}
\mathrm{MMD}
=
\frac{1}{|\mathcal{Q}|}
\sum_{Q\in\mathcal{Q}}
\min_{\hat{Q}\in\hat{\mathcal{Q}}}
\mathrm{CD}(\hat{Q},Q),
\qquad
\mathrm{COV}
=
\frac{\big|\big\{\,\arg\min_{Q\in\mathcal{Q}}\mathrm{CD}(\hat{Q},Q)\;:\;\hat{Q}\in\hat{\mathcal{Q}}\,\big\}\big|}{|\mathcal{Q}|}.
\end{equation}
COV is the fraction of ground-truth objects that are selected as the nearest neighbor of at least one generated object. We report MMD in units of \(10^{-3}\) and COV as a percentage.

\paragraph{P-FID.}
We compute the Fr\'echet distance between the distributions of generated and ground-truth objects in the feature space of the point-cloud feature extractor released with Point-E~\citep{nichol2022point}.

\paragraph{ULIP and Uni3D.}
We measure cross-modal consistency using ULIP~\citep{xue2024ulip} and Uni3D~\citep{zhou2024uni3d}. For each object, we compute the cosine similarity between the embedding of the generated point cloud and that of the corresponding input image crop used to condition its generation.

\section{Additional Results} \label{Appendix:additional-results}
\subsection{Quantitative Results}
\paragraph{Evaluation with Estimated Depth.}
To assess sensitivity to depth estimation errors, we replace the ground-truth depth maps used in the main Toys4K~\citep{stojanov2021using} comparison with predictions from MoGe3~\citep{kong2026moge}. All methods receive the same estimated depth maps following our evaluation protocol. As shown in Tab.~\ref{tab:estimated_depth_toys4k_comparison}, \ours{} retains the best performance among the compared methods across all reconstruction and physical stability metrics. In particular, its lower penetration depth (PD) and mean displacement (\(D_{\text{mean}}\)) indicate less interpenetration and greater stability during simulation. It also achieves the best MMD, COV, and P-FID, although SAM-3D obtains higher Uni3D and ULIP scores.

\paragraph{Evaluation with VLM-Inferred Relations.}
We evaluate the effect of replacing ground-truth physical relation annotations with relations inferred by a VLM~\citep{bai2025qwen3} on Toys4K~\citep{stojanov2021using}. This comparison examines whether automatically inferred relations provide sufficient guidance for interaction-conditioned generation. As shown in Tab.~\ref{tab:vlm-toys4k_comparison}, the two settings yield similar reconstruction and generative quality. Scene-level and object-level reconstruction accuracy remain close, with only small changes in geometric and pose errors. Generative quality follows a similar pattern, with modest increases in MMD and P-FID and comparable coverage and semantic similarity scores. These results suggest that VLM-inferred relations can replace ground-truth annotations at inference time with only marginal changes in reconstruction and generative quality.

\begin{table*}[t]
    \centering
    \caption{
        \textbf{Quantitative comparison on Toys4K~\citep{stojanov2021using} with estimated depth.} We compare \ours{} with baselines using depth maps estimated by MoGe3~\citep{kong2026moge}. All methods requiring depth use the same depth maps. \(\dagger\) denotes methods whose generated objects are aligned to the scene using FoundationPose~\citep{wen2024foundationpose}.
    }
    \label{tab:estimated_depth_toys4k_comparison}

    \begingroup
    \fontsize{6}{7}\selectfont
    \setlength{\tabcolsep}{2pt}
    \renewcommand{\arraystretch}{1.15}

    \resizebox{\textwidth}{!}{%
    % Method (1) + Reconstruction (6) + Generation (5)
    % + Physical Plausibility (3) = 15 columns
    \begin{tabular}{@{}l*{14}{c}@{}}
        \toprule
        %
        % Header 1: Metric groups
        \multirow{2}{*}{\textbf{Method}}
        & \multicolumn{6}{c}{\textbf{Reconstruction Quality}}
        & \multicolumn{5}{c}{\textbf{Generation Quality}}
        & \multicolumn{3}{c}{\textbf{Physical Stability}} \\
        \cmidrule(lr){2-7}
        \cmidrule(lr){8-12}
        \cmidrule(lr){13-15}
        %
        % Header 2: Individual metrics
        % Reconstruction Quality
        & CD-S~\(\downarrow\)
        & CD-O~\(\downarrow\)
        & F1-S~\(\uparrow\)
        & F1-O~\(\uparrow\)
        & IoU-B~\(\uparrow\)
        & ICP-Rot~\(\downarrow\)
        % Generation Quality
        & MMD~\(\downarrow\)
        & COV~\(\uparrow\)
        & P-FID~\(\downarrow\)
        & Uni3D~\(\uparrow\)
        & ULIP~\(\uparrow\)
        % Physical Plausibility
        & PD~\(\downarrow\)
        & \(D_{\mathrm{mean}}\,\downarrow\)
        & \(E_{\mathrm{peak}}\,\downarrow\) \\
        \midrule
        \multicolumn{15}{c}{
            \textbf{\textit{Scene Generation}}
        } \\
        \midrule
        SAM-3D
        & 24.80 & \underline{1.73} & 0.2495 & \underline{0.6250} & 0.3861 & 21.27 & \underline{2.872} & \underline{70.46} & 5.245 & \textbf{0.5981} & \textbf{0.6292}
        & 1.203 & 194.09 & 0.8269 \\             % Physical
        %
        % PixARMesh
        % & -- & -- & -- & -- & -- & --
        % & -- & -- & -- & -- & --
        % & -- & -- & -- \\
        %
        ShapeR
        & \underline{10.54} & 2.28 & \underline{0.4955} & 0.5241 & \underline{0.4841} & \underline{19.42} & 3.676 & 64.76 & 16.64 & 0.4453 & 0.4831
        & 0.8408 & 129.81 & 0.5989 \\
        WorldSculpt
        & 12.53 & 2.11 & 0.3736 & 0.5818 & 0.4593 & 19.48 & 3.113 & 69.95 & \underline{4.941} & 0.5382 & 0.5683
        & \underline{0.2280} & 135.1 & 0.6655 \\
        \midrule
        \multicolumn{15}{c}{
            \textbf{\textit{Amodal Generation + Pose Estimation}}%
        } \\
        \midrule
        Amodal3R\( ^{\dagger}\)
        & 15.02 & 2.32 & 0.3756 & 0.5438 & 0.3875 & 27.27
        & 3.468 & 64.89 & 7.102 & 0.5368 & 0.5692
        & 0.5156 & \underline{124.2} & \underline{0.6172} \\
        GENA3D\( ^{\dagger}\)
        & 18.28 & 2.81 & 0.3247 & 0.4996 & 0.3521 & 31.66
        & 4.107 & 60.80 & 8.599 & 0.4704 & 0.4977
        & 0.8366 & 132.1 & 0.6417 \\
        \midrule
        \textbf{Ours}
        & \textbf{7.56} & \textbf{1.58} & \textbf{0.6457} & \textbf{0.6679} & \textbf{0.6522} & \textbf{12.7}
        & \textbf{2.462} & \textbf{75.58} & \textbf{2.518} & \underline{0.5795} & \underline{0.6188}
        & \textbf{0.0047} & \textbf{50.49} & \textbf{0.2208} \\
        \bottomrule
    \end{tabular}%
    }
    \endgroup
\end{table*}

\begin{table*}[t]
    \centering
    \caption{
        \textbf{Quantitative comparison on Toys4K~\citep{stojanov2021using} with VLM-based reasoning.} We compare results using ground-truth physical relation annotations with those using relations inferred by a VLM~\citep{bai2025qwen3}. \ours{} maintains comparable performance with VLM-inferred relations.
    }
    \label{tab:vlm-toys4k_comparison}

    \begingroup
    \fontsize{6}{7}\selectfont
    \setlength{\tabcolsep}{2pt}
    \renewcommand{\arraystretch}{1.15}

    \resizebox{\textwidth}{!}{%
    % Method (1) + Reconstruction (6) + Generation (5)
    % + Physical Plausibility (3) = 15 columns
    \begin{tabular}{@{}l*{14}{c}@{}}
        \toprule
        %
        % Header 1: Metric groups
        \multirow{2}{*}{\textbf{Method}}
        & \multicolumn{6}{c}{\textbf{Reconstruction Quality}}
        & \multicolumn{5}{c}{\textbf{Generation Quality}}
        & \multicolumn{3}{c}{\textbf{Physical Stability}} \\
        \cmidrule(lr){2-7}
        \cmidrule(lr){8-12}
        \cmidrule(lr){13-15}
        %
        % Header 2: Individual metrics
        % Reconstruction Quality
        & CD-S~\(\downarrow\)
        & CD-O~\(\downarrow\)
        & F1-S~\(\uparrow\)
        & F1-O~\(\uparrow\)
        & IoU-B~\(\uparrow\)
        & ICP-Rot~\(\downarrow\)
        % Generation Quality
        & MMD~\(\downarrow\)
        & COV~\(\uparrow\)
        & P-FID~\(\downarrow\)
        & Uni3D~\(\uparrow\)
        & ULIP~\(\uparrow\)
        % Physical Plausibility
        & PD~\(\downarrow\)
        & \(D_{\mathrm{mean}}\,\downarrow\)
        & \(E_{\mathrm{peak}}\,\downarrow\) \\
        \midrule
        Ours (GT)
        & \textbf{3.68} & \textbf{0.97} & \textbf{0.8407} & \textbf{0.8084} & \textbf{0.7918} & \textbf{5.29}
        & \textbf{1.501} & 83.11 & \textbf{1.599} & 0.6884 & 0.7182
        & \textbf{0.0036} & \textbf{38.91} & \textbf{0.1721} \\             % Physical
        %
        % PixARMesh
        % & -- & -- & -- & -- & -- & --
        % & -- & -- & -- & -- & --
        % & -- & -- & -- \\
        %
        Ours (VLM)
        & 3.76 & 0.98 & 0.8399 & 0.8063 & 0.7905 & 5.37 & 1.538 & \textbf{83.38} & 1.651 & \textbf{0.6891} & \textbf{0.7195}
        & 0.0038 & 41.25 & 0.1780 \\
        \bottomrule
    \end{tabular}%
    }
    \endgroup
\end{table*}

\begin{figure}[t]
    \centering
    \includegraphics[width=\linewidth]{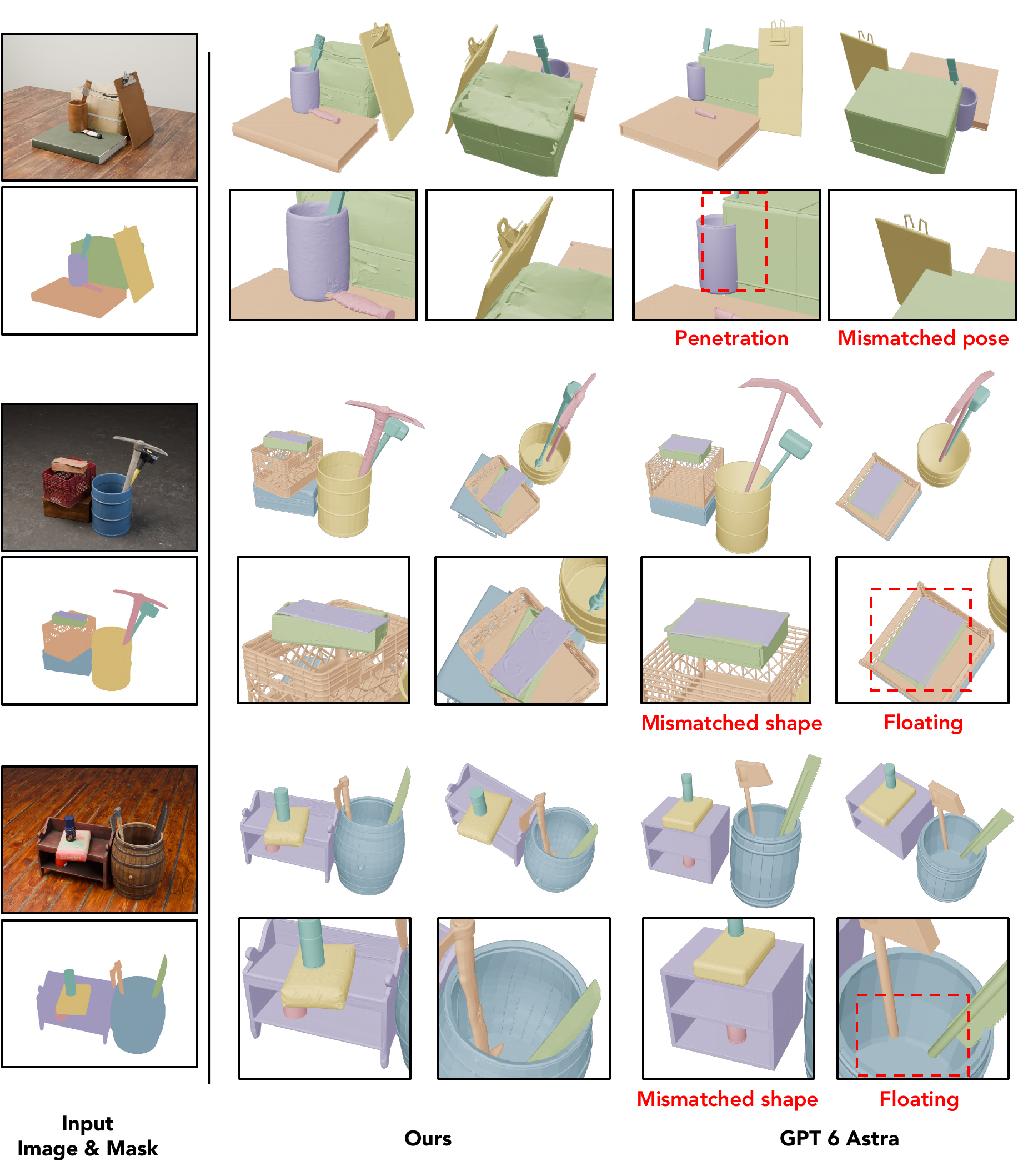}
    \caption{\textbf{Qualitative comparison with agentic 3D generation.}
    We compare \ours{} with an LLM agent-based approach to 3D mesh generation~\citep{openai2026gpt6astra}. Although this approach can produce broadly similar shapes, it often fails to faithfully recover the object geometry depicted in the input image and struggles to generate physically consistent scene layouts.
    }
    \label{figure-comparison-astra}
\end{figure}

\subsection{Qualitative Results}
\begin{figure*}[!t]
    \includegraphics[width=\linewidth]{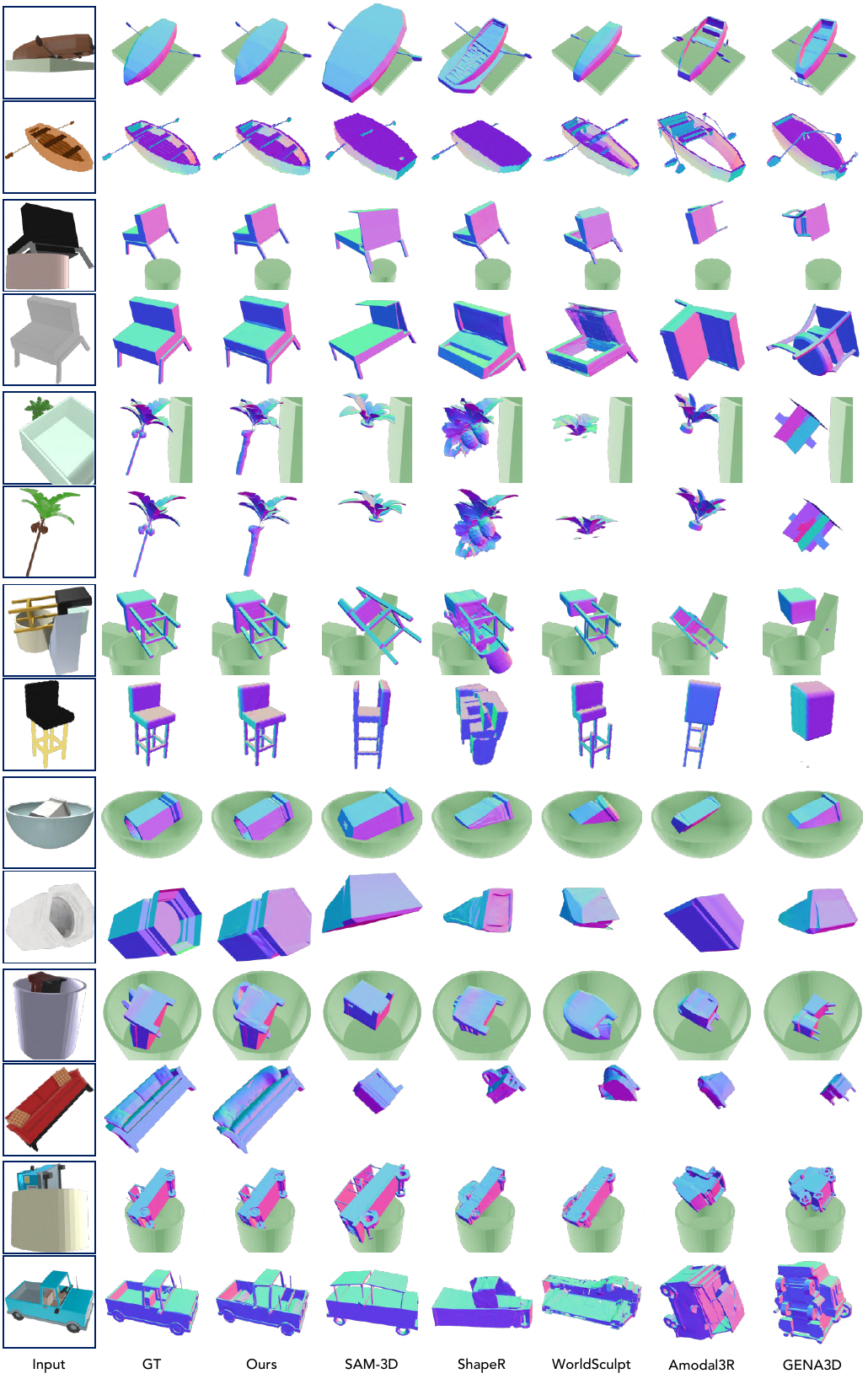}
    \caption{\textbf{Qualitative comparison with baselines on Toys4K~\citep{stojanov2021using}}.}
    \label{fig:ablation-more-qual-1}
\end{figure*}

\begin{figure*}[!t]
    \includegraphics[width=\linewidth]{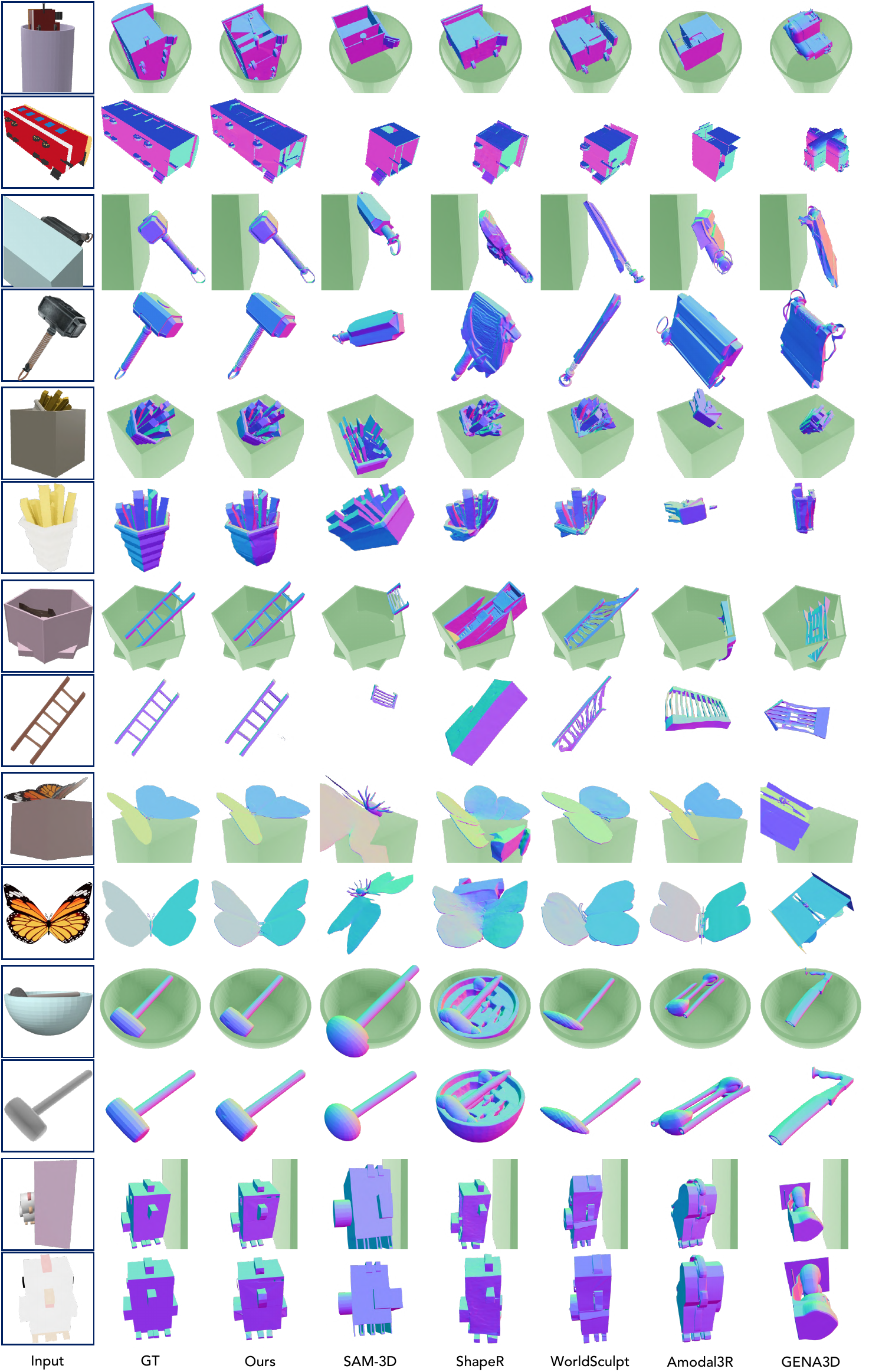}
    \caption{\textbf{Qualitative comparison with baselines on Toys4K~\citep{stojanov2021using}}.}
    \label{fig:ablation-more-qual-2}
\end{figure*}

\begin{figure*}[!t]
    \includegraphics[width=\linewidth]{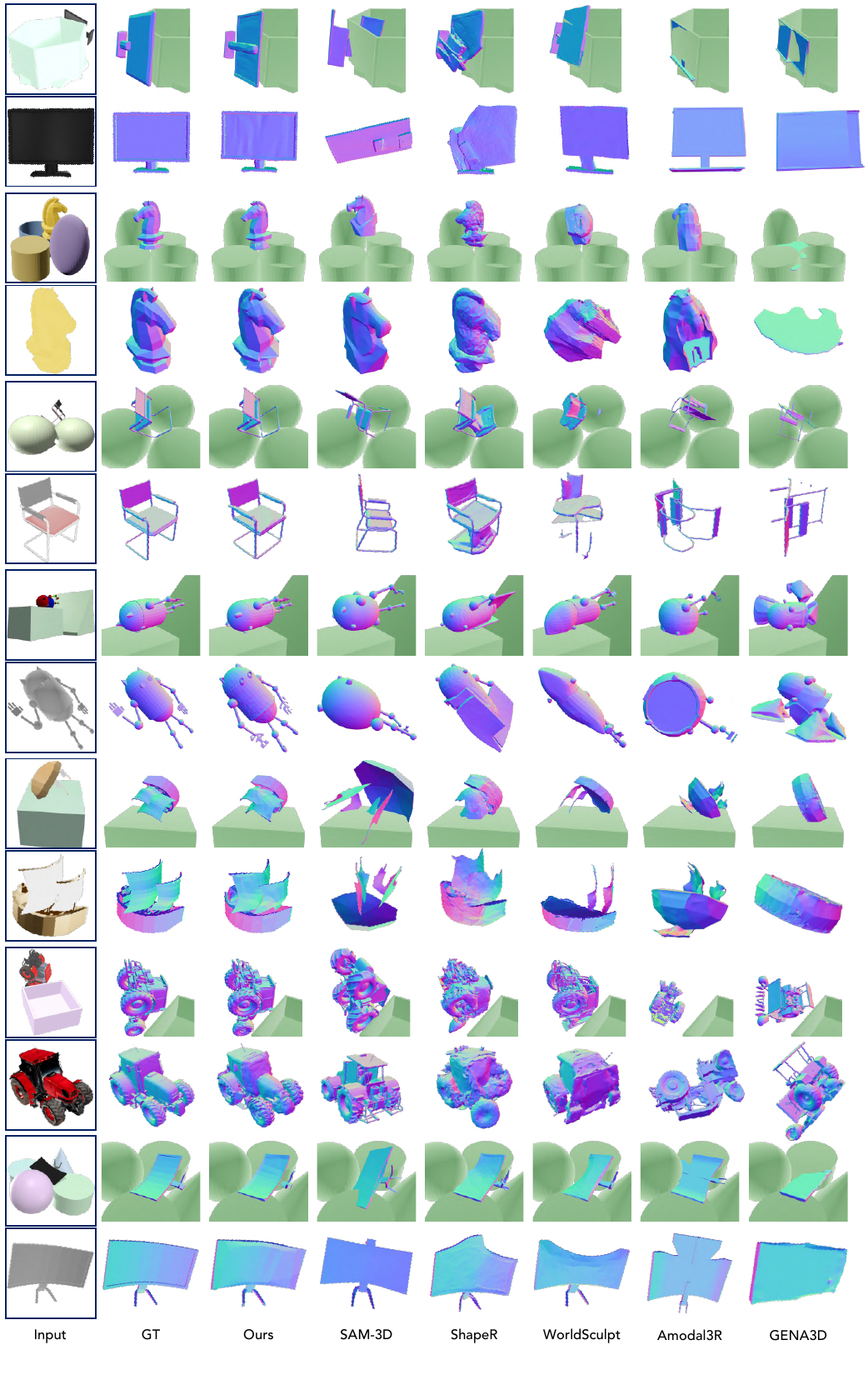}
    \caption{\textbf{Qualitative comparison with baselines on Toys4K~\citep{stojanov2021using}}.}
    \label{fig:ablation-more-qual-3}
\end{figure*}

We provide additional qualitative comparisons on Toys4K in Figs.~\ref{fig:ablation-more-qual-1}--\ref{fig:ablation-more-qual-3}. The examples include objects with thin structures and elongated parts that are partially occluded by surrounding primitives. For each example, we show both the reconstruction in scene space and the individual object from a common viewpoint, allowing shape completion and scene placement to be examined separately. By comparison, several baselines omit occluded parts, distort object proportions, or recover poses that differ from the ground truth. These examples complement the quantitative results by illustrating the benefit of considering neighboring objects when recovering both complete shapes and their spatial relationships within the scene.

\subsection{Comparison with Agentic Scene Generation}
\label{Appendix:comparison-astra}
We compare \ours{} with an agentic scene generation approach based on a frontier large language model~\citep{openai2026gpt6astra}. As shown in Fig.~\ref{figure-comparison-astra}, \ours{} reconstructs object shapes and poses that more faithfully match the input image and depth. Although the agent-based approach produces broadly similar shapes and poses, it struggles to reproduce the scene configuration depicted in the image. We attribute these discrepancies in part to its reliance on constructing CAD meshes through generated scripts and 3D graphics tools, which can limit the recovery of fine, image-specific geometry. Moreover, the agent estimates object poses through spatial reasoning from the image but struggles to place objects precisely. The resulting scenes exhibit physically invalid configurations, including penetration (top row) and unsupported floating (middle and bottom rows). These results suggest that dedicated 3D generative models are better suited to recovering image-specific object geometry in this setting. They further suggest that explicitly conditioning shape and pose generation on neighboring geometry and physical relationships can yield more physically consistent scenes than relying on spatial reasoning for object placement alone. We provide the generation prompt in Prompt~\ref{Appendix_prompt_astra}.
\section{Discussion and Limitations}
Our results suggest that physical interactions can serve as cues for recovering object shapes and poses, rather than only as criteria for evaluating a completed scene. By incorporating neighboring geometry and physical relations into generation, \ours{} uses surrounding context to guide reconstruction even when interacting regions are occluded. However, these conditions provide learned guidance rather than hard physical constraints, and therefore do not guarantee physically valid configurations in every case. Our inference pipeline relies on off-the-shelf models for depth and camera estimation, object segmentation, and physical relation reasoning. Although these modules enable reconstruction from a single RGB image, their errors can affect the conditions used for generation. Inaccurate depth or segmentation can distort the object grid and projected features, while incorrect physical relations can lead to inappropriate interaction conditions or generation orders. Since previously generated objects provide context for subsequent objects, these errors may propagate through the autoregressive process. Reducing these dependencies and accounting for uncertainty in the estimated conditions are directions for future work.

\clearpage

\clearpage
\section{Prompt Examples}

\begin{promptbox}[label={Appendix_prompt_A}]{Prompt used for physical dependency reasoning}
You are a careful 3D scene reasoning assistant.
You infer a physical dependency hierarchy: directed load-bearing edges and undirected contact edges among scene objects, including the ground plane.
This hierarchy is used for sequential physical assembly, layered from the ground up.
Each edge must be labeled with a pairwise physical relation from R = {stack, lean, contain, touch}.
stack, lean, and contain are directed. touch is undirected.
The generation order is determined by safely removing cycles from the original graph—by deleting edge pairs with the minimum sum of confidence—and then following an order that satisfies topological sorting.
The unique node with an in-degree of 0 is given as the root node, 'Plane'.
Every listed object must appear in your output.
Return strict JSON only. No markdown fences, no comments, no extra text.

You are given:
1. An original RGB scene image.
2. A color instance-segmentation image, where each object instance is painted in a unique flat RGB color.
3. An object inventory JSON. Each entry has a description and an instance RGB color [R, G, B].

How to use the inputs:
- Match each inventory name to its region by the instance_rgb_color in the segmentation image.
- Then inspect the same region in the scene image, using the description as semantic context.
- Use object names exactly as given. Do not rename, merge, split, or invent objects.

Task:
Infer every pairwise physical relation that is visible or physically necessary, and emit it as an edge. Then run the Optimal Topological Sort below on those edges and emit the recovered generation order as the last field.

Two kinds of edges (algorithm inputs):
- Directed edges (stack, lean, contain): object A relies on object B for physical support or stability. Output from="B", to="A", i.e. B -> A. In the sort, B is a possible predecessor of A under a specific confidence level.
- Undirected edges (touch): real physical contact that is not load-bearing support. Output one edge whose from/to merely name the pair; the from/to order is not generation order. In the sort, each endpoint is a possible predecessor of the other.
- Do not let a touch edge reverse, cancel, or replace a stack, lean, or contain edge.

How generation order is computed (Optimal Topological Sort). Your edges must be valid inputs to this procedure:
1. Vertices N = all objects including Plane. Root = Plane.
2. possiblePredecessors[v] starts empty for every vertex v.
3. For every directed edge (u, v, confidence): add u to possiblePredecessors[v].
4. For every undirected touch edge (u, v, confidence): add v to possiblePredecessors[u] and u to possiblePredecessors[v].
5. startState = {Plane}. fullState = N. Every subset of N that contains Plane is a state. Cost of every state is infinity except cost(startState) = 0. Each state records a parent vertex or none.
6. Process states from smaller subsets to larger. Skip a state S if its cost is still infinity.
7. For each vertex v not in S: if S contains no possible predecessor of v, skip v. Otherwise deletionCost is the sum of confidences of directed edges v -> u with u already in S. Those directed edges are treated as deleted. nextState = S union {v}. If cost(S) + deletionCost is strictly cheaper than the current cost of nextState, update that cost and set parent(nextState) = v.
8. Recover the order by walking backward from fullState to startState: repeatedly prepend parent(state) and remove it from the state. Plane is generated first; the recovered sequence is the generation order of the remaining objects. Emit that full sequence, Plane excluded, as "order".

What this implies for the edges you emit:
- An object can be generated only after at least one of its possible predecessors is already generated. An object with no possible predecessor can never be generated.
- Directed cycles are allowed. If two objects mutually load-bear, emit both directed edges with honest confidences. The sort keeps the cheaper growth path and deletes the conflicting directed edges of minimum total confidence.
- Do not drop, reverse, or canonicalise a stack, lean, or contain edge just to make a DAG. Do not invent a tie-break direction for touch; it is undirected.
- Touch does participate in order eligibility: once either contacting object is generated, the other may become eligible even if no directed support between them exists. Touch does not prefer one order over the other.
- Directed confidence is the cost of violating that ordering constraint. High confidence = clearly visible load-bearing that should be expensive to delete. Low confidence = occluded or guessed, cheaper to delete if it conflicts.

Plane:
- The ground plane is a valid support object named "Plane". Never put "Plane" in "to".
- Include a Plane -> object directed edge whenever the object rests on, stands on, or is otherwise supported by the ground. Label that edge "stack" unless a more specific type applies.
- Plane is the unique root: it is the startState, never a generated dependent, and never needs a predecessor.

Relation types (use exactly these lowercase labels):
- stack: the dependent object rests on, stands on, or is piled on the supporting object. The supporter load-bears from below, with substantial footprint overlap. This is the default for on-top support, including support by Plane. Directed: supporter -> dependent.
- lean: the dependent object is not self-standing and would topple or fall without the supporting object's lateral contact. After identifying a tilted or slanted object, add this directed edge from the object or Plane it leans against.
- contain: the dependent object is inside, partially inserted into, or rim-supported by a container. Directed: container -> contained.
- touch: there is real physical contact that is none of stack, lean, or contain (for example side-by-side contact). Use touch only when contact is visible or physically necessary, not for mere proximity, overlap, or shadow. Undirected.
- Do not use "none" and do not emit an edge for incidental proximity with no physical dependence.

Priority:
- Label priority for a given pair: contain > lean > stack > touch. Pick the highest applicable label; do not emit a lower-priority label for the same contact.
- If stack, lean, or contain applies, do not label the same contact as touch.
- Prefer physically plausible, gravity-consistent load-bearing support over mere visual adjacency.
- One object may support many objects. One object may depend on many objects at once. If physically plausible, return all corresponding directed edges, including both directions when mutual load-bearing is real.
- Incidental proximity is not an edge. Symmetric touch is one undirected edge, not a pair of directed ordering constraints.

Coverage and graph constraints:
- Every object except Plane must have at least one possible predecessor, so growth from {Plane} to the full set is possible. If the object is load-bearing-supported, that predecessor must include a directed stack, lean, or contain parent. If the support source is unknown or fully occluded, add Plane -> that object with relation "stack" and low confidence.
- "confidence" is a real number in 0.0-1.0. Use a high value when the relation is clearly visible. Use a low value when the object is occluded or the relation is guessed. For directed edges, this value is the deletion cost if the sort generates the dependent object before the supporter.
- Directed 2-cycles are allowed: you may emit both A -> B and B -> A when both load-bearing directions are physically true. Do not emit both orientations of the same touch pair; touch is one undirected edge.
- Do not emit duplicate edges. Two edges are duplicates if they share the same from, to, and relation. A given directed pair (from, to) may have only one label: do not emit two edges with the same from and to even if the relations differ. For touch, emit exactly one edge per unordered pair; put the lexicographically smaller name in "from" only as a serialization convention, not as an ordering claim.

"order" rules:
- Compute it only by the Optimal Topological Sort above, using the edges in this same JSON. Do not invent a separate assembly order.
- It is the sequence recovered in step 8. Walking backward prepends each parent, so the earliest generated object ends up at the front.
- Every name in "objects" appears exactly once, and no other name appears.

Return JSON with this schema. "order" is required and must be the last key:
{
  "vertices": ["Plane", "object_01", "object_02"],
  "edges": [
    {
      "from": "supporting_or_anchoring_object_name",
      "to": "dependent_object_name",
      "relation": "stack",
      "confidence": 0.85,
      "reason": "short explanation"
    }
  ],
  "order": ["object_01", "object_02"]
}
Allowed "relation" values: "stack" | "lean" | "contain" | "touch".

\end{promptbox}

\begin{promptbox}[label={Appendix_prompt_B}]{Object Inventory Prompt}
You are a careful visual scene parser for 3D scene reconstruction.
You inventory every instance-segmented object in a scene.
Return strict JSON only. No markdown fences, no comments, no extra text.

You are given:
1. An original RGB scene image.
2. A color instance-segmentation image, where each object instance is painted in a unique flat RGB color.

Task:
Identify every distinct object instance from the unique colors in the segmentation image.

For each object:
- Use the segmentation color only to locate the matching region in the original image.
- Keep the two color sources separate. Do not confuse them:
  - "description" appearance (shape, color, material) must come from the original RGB image. Never describe an object using its segmentation-mask paint color. A brown book painted green in the segmentation image is still a brown book, not a green book.
  - "instance_rgb_color" must be the exact flat mask color from the segmentation image. Do not copy the object's real color from the original photo.
  - If a description refers to a nearby object, use that object's real color from the original image, not its mask color.
- Ignore unlabeled background (typically black, white, or near-zero colors that do not correspond to an object instance).
- Do not merge two different colors into one object, and do not split one color into two objects.
- Name objects sequentially as object_01, object_02, ... in raster order: sort unique object colors by the first occurrence of each color, top-to-bottom then left-to-right.
- Write a description that later physical-dependency reasoning can use. Pack all of the following into the single "description" string; do not add extra JSON fields:
  (a) category,
  (b) visual appearance from the original image (shape, color, material if visible),
  (c) approximate position in the scene,
  (d) occlusion.
- Set instance_rgb_color to the exact integer [R, G, B] of that object's flat color in the segmentation image. Do not guess the color from the original photo.

Output constraints:
- Cover every unique object color. Do not invent objects that have no segmentation color.
- Return JSON only.

Return JSON following the exact format of this example:
{
  "object_01": {
    "description": "A desk lamp; brown metal with a curved stem, rounded base, and conical lampshade; standing on the left side of the desk; the shade is unoccluded.",
    "instance_rgb_color": [125, 237, 238]
  },
  "object_02": {
    "description": "A book; brown, rectangular, and thick with a box-like shape; lying to the right of the lamp; the spine is partly occluded.",
    "instance_rgb_color": [46, 204, 113]
  }
}

\end{promptbox}

\begin{promptbox}[label={Appendix_prompt_astra}]{Prompt used for agentic generation}
## Task
- Reconstruct an object-level 3D scene from the provided RGB image by authoring and executing Blender Python scripts. The resulting scene must contain independently addressable meshes corresponding to the distinct physical objects depicted in the image. Use the provided RGB image and, when available, its associated depth map as observational evidence. 

## Generation mechanism
- Construct the scene directly through agent-authored procedural modeling code in Blender. Do not invoke external image-to-3D systems, pretrained geometry-generation models, or asset-retrieval services. Create geometry using explicitly specified primitives, mesh operations, curves, modifiers, and procedural materials. Execute modeling and rendering on the CPU.

## Reconstruction objectives
- Identify the visible objects and estimate their shapes, relative dimensions, orientations, and spatial relationships from the observations. Represent each semantic object as a separate mesh, combining its constituent components where appropriate. Reproduce the observed silhouettes, relative proportions, arrangement, support relationships, and material appearance. Infer plausible geometry for occluded or unobserved regions without consulting additional scene information. Treat absolute scale and unobserved structure as assumptions unless constrained by the permitted inputs. Estimate the camera configuration and lighting from the image to support visual comparison between the reconstruction and the reference.

## Iterative refinement
- Render the reconstructed scene from the estimated reference viewpoint. Compare the rendering with the input image and revise object geometry, placement, materials, or camera parameters to reduce visible discrepancies. Validate that the exported scene preserves the intended object separation and contains usable mesh geometry.

## Deliverables
- Provide an editable Blender scene, a complete scene export, individual object mesh exports, the executable generation scripts, and preview renderings. Include an object inventory and a provenance record identifying the inputs used.

\end{promptbox}

\end{document}